\documentclass[journal]{IEEEtran}
\usepackage[utf8]{inputenc}
\usepackage{multicol}
\usepackage{multirow}
\usepackage[keeplastbox]{flushend}
\usepackage{diagbox}
\usepackage{amsmath}
\usepackage{booktabs}
\usepackage{multirow}
\usepackage{verbatim}
\usepackage{array}
\usepackage{dblfloatfix}
\usepackage{fixltx2e}
\usepackage{algorithmic}
\usepackage{array}
\usepackage{url}
\usepackage{float}
\usepackage{graphicx,psfrag,epsfig}
\usepackage{caption}
\usepackage{subfigure}
\usepackage{color}
\usepackage[export]{adjustbox}
\usepackage{cite}
\usepackage{amssymb}
\usepackage[ruled]{algorithm2e}

\begin{document}
\title{SRCNet: Seminal Representation Collaborative Network for Marine Oil Spill Segmentation}          
%\author{****}
%\date{December 2019}
\author{Fang Chen, Heiko Balzter, Peng Ren and Huiyu Zhou 
        % <-this % stops a space    
%\author{Michael~Shell,~\IEEEmembership{Member,~IEEE,}
	%John~Doe,~\IEEEmembership{Fellow,~OSA,}
	%and~Jane~Doe,~\IEEEmembership{Life~Fellow,~IEEE}% <-this % stops a space
\thanks{Fang Chen and Huiyu Zhou are with School of Computing  
and Mathematical Sciences, University of Leicester, Leicester LE1 7RH, United Kingdom (H. Zhou is the corresponding author. E-mail:  hz143@leicester.ac.uk).}% <-this % stops a space
\thanks{Heiko Balzter is Director of the Institute for Environmental Futures, National Centre for Earth Observation and School of Geography, Geology and Environment, Space Park Leicester, University of Leicester, 92 Corporation Road, Leicester, LE4 5SP, United Kingdom.}% <-this % stops a space     
\thanks{Peng Ren is with College of Oceanography and Space Informatics, China University of Petroleum (East China), Qingdao 266580, China.}}

% The paper headers
%\markboth{IEEE TRANSACTIONS ON GEOSCIENCE AND REMOTE SENSING}%
%{Shell \MakeLowercase{\textit{et al.}}: Bare Demo of IEEEtran.cls for IEEE Journals}

\maketitle

\begin{abstract} 
Effective oil spill segmentation in Synthetic Aperture Radar (SAR) images is critical for marine oil pollution cleanup, and proper image representation is helpful for accurate image segmentation. In this paper, we propose an effective oil spill image segmentation network named SRCNet by leveraging SAR image representation and the training for oil spill segmentation simultaneously. Specifically, our proposed segmentation network is constructed with a pair of deep neural nets with the collaboration of the seminal representation that describes SAR images, where one deep neural net is the generative net which strives to produce oil spill segmentation maps, and the other is the discriminative net which trys its best to distinguish between the produced and the true segmentations, and they thus built a two-player game. Particularly, the seminal representation exploited in our proposed SRCNet originates from SAR imagery, modelling with the internal characteristics of SAR images. Thus, in the training process, the collaborated seminal representation empowers the mapped generative net to produce accurate oil spill segmentation maps efficiently with small amount of training data, promoting the discriminative net reaching its optimal solution at a fast speed. Therefore, our proposed SRCNet operates effective oil spill segmentation in an economical and efficient manner. Additionally, to increase the segmentation capability of the proposed segmentation network in terms of accurately delineating oil spill details in SAR images, a regularisation term that penalises the segmentation loss is devised. This encourages our proposed SRCNet for accurately segmenting oil spill areas from SAR images. Empirical experimental evaluations from different metrics validate the effectiveness of our proposed SRCNet for oil spill image segmentation.
\end{abstract}
\begin{IEEEkeywords}
 SAR image representation, training for image segmentation, collaborative efficient learning, oil spill segmentation
\end{IEEEkeywords}

\section{introduction}
\IEEEPARstart{M}{arine} oil spills are the release of a liquid petroleum hydrocarbon or crude oil from drilling rigs, tankers or offshore platforms into the ocean waters, and the spilled oil presents heavy pollution to the marine environment \cite{zhang2019marine}. Generally, oil pollution at sea has bad effect for shellfish, seabirds, mammals and other organisms \cite{hjermann2007fish}. For example, oil spills penetrate into the fur of mammals and plumage structure of birds, making them more vulnerable to changing oceanographic factors \cite{farrington2014oil}. Thus, oil spills have disastrous consequences for society and the marine environment, especially the marine ecosystem \cite{kingston2002long} \cite{krupp2008impact}. Besides, since oil spills on the ocean surface can spread hundreds of nautical miles in a thin oil slick, they thus are generally much more damaging than those on land. Therefore, conducting timely damage assessment and spread control of oil spills is necessary for marine environmental protection \cite{soukissian2016satellite} \cite{li2017wind}. Satellite-based synthetic aperture radar (SAR) is regarded as a powerful spaceborne sensor for detecting marine oil spills \cite{solberg2007oil}, because of it has the capabilities of remotely observing environmental and natural targets on the earth surface in day and night time independence weather conditions with wide coverage \cite {brekke2005oil} \cite{velotto2016first}. In addition, for image processing, it is recognised that the processing performance is correlated to the image data representation that describes image formation \cite{yin2014modified}. Therefore, to conduct effective oil spill image segmentation for oil pollution cleanup, we develop an automatic and effective segmentation strategy by leveraging SAR image representation and the training for oil spill segmentation simultaneously for accurately segmenting oil spills in SAR images.  

Specifically, in the operation of oil spill SAR image data representation, researchers from geoscience and remote sensing community mainly devote to investigating the description of oil spill SAR images with respect to physical characteristics to develop strategies for analysing oil spills in SAR image data. The physical description of SAR images corresponds to the interpretation of electromagnetic wave illumination on the ocean surface. Particularly, the polarimetric characteristics that characterising oil spills in SAR images have been delicately investigated, and the representative researches from this perspective include those by Salberg et al. \cite{salberg2014oil},  Migliaccio et al. \cite{buono2016polarimetric} \cite{velotto2011dual}, Espeset et al. \cite{espeseth2017analysis}, Minchew et al. \cite{minchew2012polarimetric} \cite{collins2015use}. They conducted the studies based on the polarimetric properties of SAR image data correlating to SAR imagery for oil spill observations. Thus, the polarimetry based methodologies facilitate effective oil spill characterisation in SAR images. This enables the basic image processing techniques such as graph theory \cite{gemme2018automatic} and thresholding \cite{lupidi2017fast} methods can be implemented to detect oil spills in SAR images with enhanced representations. 

On the other hand, in the implementation of image processing, researches from machine learning community concentrate on constructing segmentation strategies that are capable of performing effective oil spill segmentation in SAR images. Generally, most state-of-the-art segmentation techniques in this field are developed with respect to energy minimisation, where an energy functional is constructed with the objective of measuring the segmentation operation with respect to similarity and fitness of oil spills. Particularly, Marques et al. \cite{marques2011sar} constructed SAR image segmentation energy functional by exploiting the model representation of images in different regions. Xu et al. \cite{xu2017level} operated oil spill SAR image segmentation with the construction of the segmentation energy functional using an arbitrary pixel within one marine oil spill region as initialisation. Xia et al. \cite{xia2015meaningful} performed oil spill SAR image segmentation with the integration of the non-local interactions between patch pairs to formulate the segmentation functional. Chen et al. \cite{chen2018segmenting} constructed the segmentation energy functional with the exploitation of the alternating direction method of multipliers to formulate the segmentation energy functional, by which the constructed segmentation method achieves oil spill segmentation from blurry SAR images. Jing et al. \cite{jing2011novel} utilised global minimisation active contour model to define the segmentation energy functional, and this performs oil spill detection without local minima. Moreover, to address the segmentation of oil spill SAR images in terms of precisely preserving oil spill edges in the segmentation results, Chen et al \cite{chen2017level} presented an edge sensitive segmentation scheme with self-guided filtering incorporated to construct the segmentation energy functional.
These mentioned strategies operate oil spill image segmentation by fitting region information with active contour models, and the segmentation operated by manually labelling an initial region for starting the segmentation. Therefore, the segmentation output is correlated to the captured initialisation region and an proper initialisation region given contributes to a more accurate segmentation \cite{ren2018energy}. However, in SAR observation for maritime scenes, turbulence and wind blowing on the ocean surface results in oil spills in SAR images exhibit various areas with irregular shapes \cite{bertacca2005farima}. This makes difficulty for accurately delineating oil spills in SAR images and thus manual initialisation labelling is hard to capture reliable region for stable and accurate oil spill segmentation. In this scenario, in the implementation of oil spill image segmentation, to render effective segmentation, we aim to develop an automatic and efficient methodology for oil spill image segmentation.   

Particularly, to achieve effective oil spill image segmentation, different from the methodologies that either focus on exploring image physical characterisation or formulating segmentation energy functional with manual intervention for performing segmentation, in this work, we comprehensively consider SAR image formation and oil spill segmentation to construct a seminal representation collaborative segmentation network. This proposed segmentation technique leverages SAR image representation and the training for oil spill segmentation simultaneously and thus enables accurate oil spill segmentation. Specifically, in our proposed segmentation network, to operate effective oil spill image segmentation, two deep neural nets are structured, with one is a generative neural net, which is trained to produce oil spill segmentation maps by drawing the samplers from the given seminal representation of SAR images, and the other is a discriminative net, which is trained to distinguish the generated segmentation from the real oil spill segmentation as much as possible. This beneficial for improving the capability of the generative net in terms of generating accurate segmentation maps, and the training for these two deep neural nets is conducted simultaneously, with the training achieves convergence with the generated segmentations are as accurate as the ground-truth segmentations. Additionally, in the operation of image segmentation, many standard methods operate image mapping from input to output using networks in an encoder-decoder form \cite{hinton2006reducing}. Such kind of network requires the input information passes through a series of layers including the bottleneck layer, and this normally results in bottleneck problem. In the operation of deep learning based image segmentation, this problem is addressed through the utilise of skip connections, which build the correlations between the feature maps from the final layers and the earlier layers, making model combines information from both deep and shallow layers to produce accurate segmentations \cite{minaee2021image}. Thus, to produce accurate oil spill segmentations, we construct our proposed segmentation network with skip connection technique exploited. This structurally updates the generative net in a symmetric form with the mirrored deep neural layers are connected, enabling information shuttled across the net directly, and thus provides a way for the generative net to circumvent the bottleneck problem to render more accurate oil spill image segmentation. Furthermore, in the training for oil spill image segmentation, to increase the segmentation capability of our proposed segmentation network in terms of encouraging the generated segmentation approaches the ground-truth segmentation, a squared loss term is formulated, which regularizes the segmentation for accurately delineating oil spill areas, and thereby enhances the segmentation performance of our proposed SRCNet with fast speed. The main contributions of our proposed oil spill image segmentation method are summarised as follows:

$\bullet$ We propose a novel marine oil spill segmentation network named SRCNet, which is structured with two deep neural nets with the collaboration of seminal representation of SAR images. In our proposed SRCNet, the mirrored layers are connected symmetrically. This enables direct information shuttle in the network to conduct effective oil spill segmentation.

$\bullet$ We operate the training for these two deep neural nets simultaneously, and the collaborated seminal representation contributes to conduct efficient learning for accurate oil spill segmentation without requiring large amount of qualified training image data. This provides an economical way for implementing oil spill SAR image segmentation. 

$\bullet$ We devise a squared loss term into our segmentation model to enable the segmentation clearly dealinate oil spill areas in SAR images. This enhances the segmentation capability of our proposed SRCNet for accurately segmenting oil spill areas with irregular shapes.

Extensive experimental evaluations from different metrics validate the effectiveness of our proposed method for oil spill segmentation over different sources of SAR images.

\section{Related Work}
In the implementation of oil spill image segmentation, a segmentation technique that is capable of performing effective and automatic segmentation is pursued. In this regard, in our segmentation work, we construct our proposed oil spill image segmentation network by taking the advantage of recent advances of deep neural networks in terms of training for image segmentation by consideration image representation. Specifically, in the application of deep neural networks for image segmentation, a striking point is the mapping of high-dimensional input to a class label with the involvement of discriminative models \cite{krizhevsky2017imagenet}. Therefore, to present the detailed procedures of constructing our proposed segmentation network, we here briefly review the deep neural network which  performs image segmentation by pairing the generative net that is trained to produce segmentation maps with a discriminative model. Specifically, to describe the segmentation operation, let define ${p}_{data}(\textbf x)$ be the distribution of input data, $p_{\textbf z}(\textbf z)$ be the prior distribution of input noise variables $\textbf z$, $p_{gen}(\textbf x)$ be the learned distribution of the generative net over data $\textbf x$, $D(\textbf x)$ be the probability that the discriminative net obtaining $\textbf x$ from data rather than $p_{gen}$, and $G(\textbf z)$ represents the probability of mapping the input variables $\textbf z$ to data space. The generative and the discriminative nets build the value function shown as follows:
\begin{equation}
\begin{split}
  \mathcal{V} \big(G,D\big)&\!\!=\!\!\!\int\!\! p_{data}(\textbf x)\log\!\big(D(\textbf x)\!\big)dx\!+\!\!\!\int \!\! p_{\textbf z}(\textbf z)\log\!\big(1\!\!-\!D(G(\textbf z)\!)\!\big)dz \\
  &\!\!=\!\!\int p_{data}(\textbf x)\log\big(D(\textbf x)\big)+p_{gen}(\textbf x)\log\big(1-D(\textbf x)\big)dx
  \label{value function between generator and discriminator}
\end{split}
\end{equation} 
In the operation for image segmentation, the objective is to learn the distribution $p_{gen}(\textbf x)$ that enables producing accurate segmentation maps. For clarity in terms of representing the learning operation, we firstly rewrite Eq. (\ref{value function between generator and discriminator}) as follows:
\begin{equation}
\begin{split}
  \mathcal{V} \big(G,D\big)&\!\!=\!\!\!\int\!\! p_{data}(\textbf x)\log\!\big(D(\textbf x)\!\big)dx\!+\!\!\!\int\!\! p_{\textbf z}(\textbf z)\log\!\big(1\!\!-\!D(G(\textbf z)\!)\!\big)dz\\
  &\!\!=\mathbb{E}_{\textbf x\sim p_{data}(\textbf x)}\!\big[\!\log D(\textbf x)\!\big]\!+\!\mathbb{E}_{\textbf z\sim p_{\textbf z}(\textbf z)}\!\big[\!\log(1\!-\!D(G(\textbf z)\!)\!)\!\big]
  \label{value function in expectation}
\end{split}
\end{equation}
This equation represents the value function with the expectation form where $\log D(\textbf x)$ and $\log(1\!-\!D(G(\textbf z)\!)\!)$ are computed over $p_{data}(\textbf x)$ and $p_{\textbf z}(\textbf z)$, respectively. In the training process, the objective of the generator is to produce outputs that as approach as the real data,  while the discriminator tries its best to distinguish between samples that come from data and the generator outputs. In other words, the generator is trained to generate accurate segmentation outputs by minimising $\log(1\!-\!D(G(\textbf z)\!)\!)$, while the discriminator is trained to maximise the probability of assigning correct label to samples from both real image data and the generator. They thereby build a two-player mini-max game with the value function $\mathcal{V} \big(G,D\big)$ is represented as follows:
\begin{equation}
\begin{split}
   \min_{G} \max_{D} \mathcal{V} \big(G,D\big)=&\mathbb{E}_{\textbf x\sim p_{data}(\textbf x)}\big[\log D(\textbf x)\big]+\mathbb{E}_{\textbf z\sim p_{\textbf z}(\textbf z)}\\
   &\big[\log (1-D(G(\textbf z)))\big]  
   \label{optimisation of value function}
\end{split}
\end{equation}
Examining this formulation, it is obvious that the inner loop is to maximise $D$ in terms of discriminating samples, and this is achieved by optimising the value function with respect to the $D$ component. From the representation $ p_{data}(\textbf x)\log\big(D(\textbf x)\big)+p_{gen}(\textbf x)\log\big(1-D(\textbf x)\big)$ in Eq. (\ref{value function between generator and discriminator}), we derive the computation regarding to $D(\textbf x)$ shown as follows:
\begin{equation}
\begin{split}
    &p_{data}(\textbf x)\frac{1}{D(\textbf x)}-p_{gen}(\textbf x)\frac{1}{\big(1-D(\textbf x)\big)}\\
    &=\frac{p_{data}(\textbf x)-\big(p_{data}(\textbf x)+p_{gen}(\textbf x)\big)D(\textbf x)}{D(\textbf x)\big(1-D(\textbf x)\big)}
    \end{split}
\end{equation}
From this equation, it is easily to obtain the optimal solution for $D$ shown as follows:
\begin{equation}
  D_{G}^{*}(\textbf x)=\frac{p_{data}(\textbf x)}{p_{data}(\textbf x)+p_{gen}(\textbf x)} 
  \label{optimisation for discriminator}
\end{equation}
Eq. (\ref{optimisation for discriminator}) shows the optimal representation for $D$. In the training process, Eq. (\ref{optimisation of value function}) shows that the training criterion for $D$, given any $G$, is to maximise the quantity of the value function $\mathcal{V} \big(G,D\!\big)$. Thus, in combination with Eq. (\ref{optimisation for discriminator}), we obtain the optimal training criterion for the mini-max operation in Eq. (\ref{optimisation of value function}) can be reformulated as follows:
\begin{equation}
\begin{split}
    C(G)&=\max_{D} \mathcal{V} \big(G,D\!\big)\\
    &=\!\!\mathbb{E}_{\textbf x\sim p_{data}}\!\big[\!\log \!D^{*}(\textbf x)\!\big]\!\!+\!\!\mathbb{E}_{\textbf z\sim p_{\textbf z}}\!\big[\!\log (1\!\!-\!\!D^{*}(G(\textbf z)\!)\!)\!\big]\\
    &=\mathbb{E}_{\textbf x\sim p_{data}}\!\big[\!\log \!D^{*}(\textbf x)\!\big]\!\!+\!\!\mathbb{E}_{\textbf x\sim p_{\textbf gen}}\!\big[\!\log (1\!\!-\!\!D^{*}(\textbf x)\!)\!\big]\\
    &=\mathbb{E}_{\textbf x\sim p_{data}}\!\big[\!\log \!\frac{p_{data}(\textbf x)}{p_{data}(\textbf x)+p_{gen}(\textbf x)}\!\big]\!\!+\!\!\mathbb{E}_{\textbf x\sim p_{\textbf gen}}\\
    &\quad\big[\!\log \frac{p_{gen}(\textbf x)}{p_{data}(\textbf x)+p_{gen}(\textbf x)}\big]
   \end{split}
   \label{minimax computation}
\end{equation}
This formulation interprets the optimality in terms of implementing training for image segmentation. In practical training process, the generative net is trained to produce samples that the discriminator unable to differentiate, i.e. $p_{gen}(\textbf x)$ equals to $p_{data}(\textbf x)$. This criterion enables the minimisation for $G$ achieved. From the description of performing image segmentation with the segmentation network is constructed combing the generative model $G$ and the discriminative model $D$, and the training criterion for these two models, it is obtain that the training efficiency and the segmentation performance depend highly on the speed of the generative net in terms of generating high quality of accurate segmentation outputs.

\section{Construction Oil Spill Segmentation Network via Seminal Representation of SAR Images}
\label{the construction of oil spill segmentation network}
In the application of deep neural networks for processing intelligent tasks, one promise is to learn models that represent probability distribution functions over data encountered in the implementation tasks \cite{bengio2009learning}. Based on this and the distribution correlations between different modules involved in deep neural segmentation networks. To render effective oil spill image segmentation, in this section, we construct our proposed oil spill segmentation network with the collaboration of the seminal distribution representation that characterises oil spill SAR images. The seminal distribution representation collaborated enables the proposed segmentation network to conduct efficient learning for producing accurate oil spill segmentation.

\subsection{Seminal Representation of Marine Oil Spill SAR Images}
\label{seminal representation of oil spill SAR images}
We operate oil spill image segmentation by constructing a segmentation network that is capable of performing effective learning for generating accurate oil spill segmentation maps, and this is achieved with the collaboration of the seminal distribution representation that characterises oil spill SAR images. Thus, to present the details of construction our proposed segmentation network, we here investigate the representation of oil spill SAR images. According to the exploration of SAR imagery for observing oceanic scenes and the established distribution representation for SAR images \cite{ulaby2014microwave} \cite{oliver2004understanding}, we obtain the distribution representation that describes SAR images shown as follows:
\begin{equation}
\mathcal{P}_{sedi}(\mathcal{I}(\textbf x))=\frac{1}{K_{s}\sigma(\textbf x)}\rm exp^{(-\frac{1}{K_{s}\sigma(\textbf x)}\mathcal{I}(\textbf x))}, \mathcal{I}(\textbf x)\geq0
\label{seminal distribution of SAR images}
\end{equation}
where $I(\textbf x)$ represents SAR image vector, $K_{s}$ is the detection system constant, and $\sigma$ is the normalized radar cross section component. Particularly, in SAR imagery for depicting oil spills, the appearance of oil spill areas in SAR images is correlated to $\sigma$, the power of backscattered radar signal \cite{brekke2005oil} \cite{alpers2017oil}. Besides, in the distribution representation, the component $\sigma$ is information bearing of SAR image pixels \cite{oliver2004understanding}, and thus it carries the information of each image pixel. Therefore, Eq. (\ref{seminal distribution of SAR images}) models the seminal distribution representation that characterises oil spill SAR images, and it thereby provides internal feature information for detecting oil spills in SAR images. Thus, in the implementation of oil spill SAR image segmentation, constructing the segmentation network with the collaboration of the image seminal representation contributes to achieve effective segmentation.

\subsection{Construction of the Seminal Representation Collaborative Network for Oil Spill Segmentation}
We conduct oil spill SAR image segmentation by taking the advances of deep learning in terms of operating image segmentation with two deep neural nets that are structured to play different roles in the segmentation operation. Our proposed segmentation network performs oil spill image segmentation in an end-to-end manner and the training for these two nets are conducted simultaneously. Besides, in the implementation of SAR image segmentation using deep neural networks, a proper image representation model utilised benefits to improve segmentation performance \cite{chen2023dgnet}. Thus, to render effective oil spill image segmentation, we construct our proposed segmentation network by leveraging SAR image representation and the training for oil spill segmentation simultaneously. This enables the proposed segmentation network obtaining the internal characteristics of SAR images and therefore performing efficient learning for accurate oil spill segmentation. The seminal distribution representation that describes oil spill SAR images is shown in Section \ref{seminal representation of oil spill SAR images}, and we next give the detailed processes of formulating our proposed segmentation technique. 

\begin{figure*}[h] 
\begin{center}
%\vspace*{-28.6mm}
%\raggedright
\includegraphics[width=1.08\textwidth,height=0.47\textheight,center]{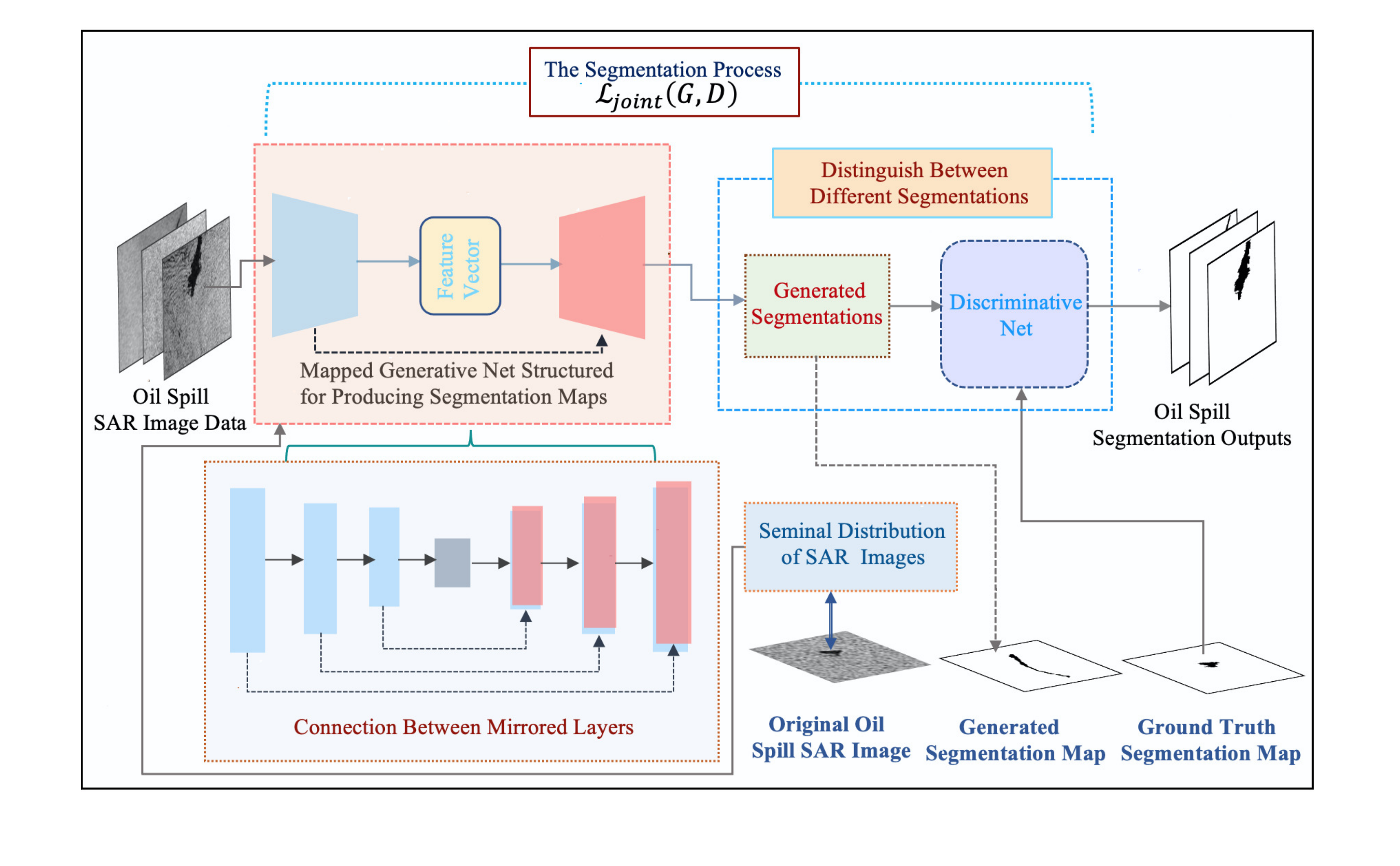}
\end{center}
\vspace*{-4.7mm}
\caption{The architecture of the proposed oil spill SAR image segmentation framework. Specifically, the proposed segmentation network is constructed with a mapped generative neural net and a discriminnative net, with the mapped generative net is structured to produce oil spill segmentation maps, and the discriminative net is trained to discriminate between the generated and the ground truth segmentations.}
\label{the proposed  oil spill segmentation framework}
\end{figure*}

Specifically, in our proposed segmentation network, one deep neural net is structured as mapped generative net, which is trained to produce oil spill segmentation maps, and the other is the discriminative net, which is trained to distinguish between the generated segmentation and the ground truth segmentation, and the training of the segmentation network reaches convergence with the generated segmentation is as accurate as the ground truth segmentation. Therefore, to render accurate oil spill image segmentation in high efficiency, we adapt our proposed segmentation network by feeding the generative net with samples drawing from the seminal distribution representation of oil spill SAR images. The seminal distribution is modelled with the physical characteristics of SAR images, providing the generative net internal image information to produce accurate oil spill segmentation maps at a fast speed. Moreover, to circumvent the bottleneck problem arises in the generative process, we construct our mapped generative net with the mirrored deep neural layers are connected to enable the desired low level image information shuttled directly across the net to achieve detailed oil spill segmentation. Thus, to formulate the interaction between these two deep neural nets for indicating the segmentation operation, let denote $\mathcal {P}_{sedi}(\textbf x)$ be the seminal distribution representation that characterises oil spill SAR images, $\mathcal {G}_{gedi}(\textbf y)$ be the distribution representation of the generated segmentation outputs, and $\mathcal{G}_{gtdi}(\textbf y)$ be the distribution representation of the ground truth segmentations. Supposing the segmentation is operated for SAR images with domain $\Omega$, and in the training for oil spill image segmentation, the loss measurement is given as follows:

\begin{equation}
\begin{split}
  \mathcal{L}_{seg} \!\big(\!G,\!D\!\big)&\!\!=\!\!\!\!\int\!\!\! \mathcal{G}_{gtdi}(\!\textbf y\!)\!\log\!\big(\mathcal{D}(\!\textbf y\!)\!\big)dy\!+\!\!\!\!\int \!\!\!\mathcal{P}_{sedi}(\!\textbf x\!)\!\log\!\big(1\!\!-\!\!D(G(\!\textbf x\!)\!)\!\big)dx \\
  &\!\!=\!\!\mathbb{E}_{\textbf y\sim \mathcal{G}_{gtdi}}\big[\!\log D(\textbf y)\!\big]\!\!+\!\mathbb{E}_{\textbf x\sim\mathcal{P}_{sedi}}[\log\big(1\!-\!D(G(\textbf x)\!)\!\big)\!\big]\\
  \label{loss function between generator and discriminator}
\end{split}
\end{equation}
This formulation builds an interactive correlation between the generative and the discriminative components, and thus operates oil spill image segmentation in an end-to-end manner. Particularly, in the representation of oil spill SAR images, the oil spills in SAR images normally exhibit various areas with irregular shapes \cite{bertacca2005farima}. This brings further challenge for accurate oil spill segmentation. To address this, in the operation of oil spill SAR image segmentation, we strive to improve the segmentation performance of our proposed segmentation network by increasing it capability in terms of accurately delineating oil spill areas in SAR images. Besides, for deep learning based segmentation network constructed with generative net, it is helpful to couple the segmentation loss with such a loss which encourages capturing more contents for accurate segmentation \cite{pathak2016context}. In this scenario, in the construction of our proposed segmentation network, we task the generative net to produce oil spill segmentation maps that approach the ground truth segmentations with a loss term that regularises more accurate oil spill segmentation is formulated, and the loss function is shown as follows: 
\begin{equation}
\mathcal {L}_{sreg}(G)= \mathbb{E}_{\textbf x\sim \mathcal{P}_{sedi},\textbf y\sim \mathcal{G}_{gedi}}[\parallel \textbf y-G(\textbf x)\parallel _{2}] 
\label{generation loss function}
\end{equation}
Clearly, this formulation measures the loss in terms of training for generating oil spill segmentation maps. Therefore, this loss function constructed in our proposed segmentation network contributing to enhance the segmentation performance. Combining Eqs.(\ref{loss function between generator and discriminator}) and (\ref{generation loss function}), we formulate the joint loss function shown as follows:
\begin{equation}
\begin{split}
  \mathcal{L}_{joint}&\!\!=\gamma_{seg}\mathcal{L}_{seg} \big(G,\!D\!\big)+ \gamma_{sreg} \mathcal {L}_{sreg}(G)\\
  &\!\!=\!\gamma_{seg}\big[\mathbb{E}_{\textbf y\sim \mathcal{G}_{gtdi}}\big[\!\log D(\textbf y)\!\big]\!\!+\!\mathbb{E}_{\textbf x\sim\mathcal{P}_{sedi}}[\log\big(1\!\!-\!\!D\\
  &(G(\textbf x)\!)\!\big)\!\big]\big]+\gamma_{sreg}\mathbb{E}_{\textbf x\sim \mathcal{P}_{sedi},\textbf y\sim \mathcal{G}_{gedi}}[\parallel \!\!\textbf y\!-\!G(\textbf x)\parallel _{2}] 
  \label{joint loss function}
\end{split}
\end{equation}
where $\gamma_{seg}$ and $\gamma_{sreg}$ are the weight coefficients that balance the loss functions for effective oil spill segmentation. Eq. (\ref{joint loss function}) formulates the objective function in our proposed segmentation network, in which the two modules $G$ and $D$ are incorporated. Thus, in the implementation of oil spill SAR image segmentation, to achieve optimal segmentation, we optimise the objective function with respect to $G$ and $D$ shown as follows:
\begin{equation}
\min_{G} \max_{D}\!\!\quad\!\!\gamma_{seg}\mathcal{L}_{seg} \big(G,\!D\!\big)+ \gamma_{sreg} \mathcal {L}_{sreg}(G)
\end{equation}
Our proposed segmentation network performs oil spill segmentation in an end-to-end manner with the discriminative net and the mapped generative net are constructed to work in an interactive manner. In the training process, these two nets are trained simultaneously to operate oil spill SAR image segmentation. The framework of the proposed segmentation network is illustrated in Fig. \ref{the proposed  oil spill segmentation framework}.

\section{Implementation Training for Oil Spill Image Segmentation}
\label{optimisation for segmentation}
We propose an oil spill segmentation network by leveraging SAR image representation and the training for oil spill segmentation simultaneously. The segmentation network is constructed with the collaboration of the image seminal representation that describes pixel-wise characteristics of SAR images, and the details of constructing the segmentation network is presented in Section \ref{the construction of oil spill segmentation network}. Specifically, in our proposed segmentation network, two deep neural nets are structured to achieve oil spill image segmentation in an interactive manner, where one is the mapped generative net which is structured to produce oil spill segmentation maps by drawing samples from the seminal distribution representation, and the other is the discriminative net which is structured to distinguish between the generated segmentation and the ground truth segmentation. Therefore, in the training for oil spill image segmentation, the mapped generative net aims to produce oil spill segmentation maps that as approach as the ground truth segmentation so as for fooling the discriminator, while the discriminative net strives to discriminate segmentations that from the mapped generative net or from the ground truth. They thus build a min-max game for conducting oil spill image segmentation, and convergence achieves with the generated segmentations are as accurate as the ground truth segmentations that makes the discriminative net unable to distinguish. To facilitate the training process, we represent the optimisation for the min-max operation as follows:

\begin{equation}
\begin{split}
&\min_{G} \max_{D}\mathcal{L}_{joint} \!\big(\!G,\!D\!\big)\!\!=\min_{G} \max_{D}\big[\gamma_{seg}\mathcal{L}_{seg} \big(G,\!D\!\big)+\!\gamma_{sreg} \\
& \mathcal {L}_{sreg}(G)\big]
  \!\!=\min_{G} \max_{D}\gamma_{seg}\big[\mathbb{E}_{\textbf y\sim \mathcal{G}_{gtdi}}\big[\!\log D(\textbf y)\!\big]\!\!+\mathbb{E}_{\textbf x\sim\mathcal{P}_{sedi}}\\
  &[\log\big(1\!\!-\!\!D(G(\textbf x)\!)\!\big)\!\big]\big]\!+\!\gamma_{sreg}\mathbb{E}_{\textbf x\sim \mathcal{P}_{sedi},\textbf y\sim \mathcal{G}_{gedi}}[\parallel \textbf y\!-\!G(\textbf x)\parallel _{2}] 
\end{split}
\label{optimisation of joint loss function}
\end{equation}
Particularly, we perform pixel wise segmentation for oil spill images to enable each pixel in image domain is accurately segmented, and thus for an input oil spill SAR image $\mathcal{I}(\textbf x)$ with $N$ datapoints in the image domain $\Omega$ (i.e. $\mathcal{I}(\textbf x)=\{\mathcal{I}(\textbf x^{(i)})\}_{i=1}^{N}$), we represent the objective function in Eq. (\ref{optimisation of joint loss function}) with respect to individual datapoint as follows:
\begin{equation}
\begin{split}
&\min_{G} \max_{D}\mathcal{L}_{joint} \big(G,D\big)=\min_{G} \max_{D}\gamma_{seg}\big[\mathbb{E}_{\textbf y^{(i)}\sim \mathcal{G}_{gtdi}}\\
&\big[\log D(\textbf y^{(i)})\big]\!+\!\mathbb{E}_{\textbf x^{(i)}\sim\mathcal{P}_{sedi}}[\log\big(1\!\!-\!\!D(G(\textbf x^{(i)})\!)\!\big)\!\big]\big]\!\!+\!\!\gamma_{sreg}\\
&\quad\mathbb{E}_{\textbf x^{(i)}\sim \mathcal{P}_{sedi},\textbf y^{(i)}\sim \mathcal{G}_{gedi}}[\parallel \textbf y^{(i)}\!-\!G(\textbf x^{(i)})\parallel _{2}] 
\end{split}
\label{optimisation of loss function}
\end{equation}
The application of deep learning for intelligent tasks is to learn models that describe the distribution representations over the employed data \cite{bengio2009learning}, and the distributions are characterised by different parameters for separate tasks. In our proposed SRCNet, two deep neural modules are constructed to play different roles, where one is the mapped generative net $G$ which is structured to produce oil spill segmentation maps, and the other is the discriminative net $D$ which is structured to discriminate the generated segmentation and the ground truth segmentation. Thus, in the implementation of oil spill SAR image segmentation with our proposed SRCNet, we exploit $\varphi_{ptg}$ and $\varphi_{ptd}$ to denote the parameters of the components $G$ and $D$, respectively. In the training process, to learn optimal distribution representation for describing the segmentation work, we optimise $\varphi_{ptg}$ and $\varphi_{ptd}$ rather than $G$ and $D$, and we thus rewritten Eq. (\ref{optimisation of loss function}) as follows:

\begin{equation}
\begin{split}
&\min_{\varphi_{ptg}} \max_{\varphi_{ptd}}\mathcal{L}_{joint} \big(G,D\big)=\min_{\varphi_{ptg}} \max_{\varphi_{ptd}}\gamma_{seg}\big[\mathbb{E}_{\textbf y^{(i)}\sim \mathcal{G}_{gtdi}}\\
&\big[\log D(\textbf y^{(i)})\big]\!+\!\mathbb{E}_{\textbf x^{(i)}\sim\mathcal{P}_{sedi}}[\log\big(1\!\!-\!\!D(G(\textbf x^{(i)})\!)\!\big)\!\big]\!\big]\!\!+\!\!\gamma_{sreg}\\
&\quad\mathbb{E}_{\textbf x^{(i)}\sim \mathcal{P}_{sedi},\textbf y^{(i)}\sim \mathcal{G}_{gedi}}[\parallel \textbf y^{(i)}\!-\!G(\textbf x^{(i)})\parallel _{2}] 
\end{split}
\label{optimisation parameters for loss function}
\end{equation}
This formulation shows that the optimisation for $\varphi_{ptd}$ and $\varphi_{ptg}$ is conducted with respect to the maximisation operation and the minimisation operation, respectively. Examining this equation, we represent the optimisation in terms of the maximisation step given as follows:
\begin{equation}
\begin{split}
&\max_{\varphi_{ptd}}\mathcal{L}_{joint} \big(G,D\big)= \max_{\varphi_{ptd}}\gamma_{seg}\big[\mathbb{E}_{\textbf y^{(i)}\sim \mathcal{G}_{gtdi}}\\
&\big[\log D(\textbf y^{(i)})\big]
\!+\!\mathbb{E}_{\textbf x^{(i)}\sim\mathcal{P}_{sedi}}[\log\big(1\!\!-\!\!D(G(\textbf x^{(i)})))\big]\big]
\end{split}
\label{optimisation with respect to discriminator}
\end{equation}
and the minimisation step given as follows:
\begin{equation}
\begin{split}
&\min_{\varphi_{ptg}}\mathcal{L}_{joint} \big(G,D\big)\!\!= \min_{\varphi_{ptg}}\gamma_{seg}\mathbb{E}_{\textbf x^{(i)}\sim\mathcal{P}_{sedi}}[\log\big(1\!\!-\!\!D(G\\
&(\textbf x^{(i)})))\big]\!\!+\!\!\gamma_{sreg}\mathbb{E}_{\textbf x^{(i)}\sim \mathcal{P}_{sedi},\textbf y^{(i)}\sim \mathcal{G}_{gedi}}[\parallel \textbf y^{(i)}\!-\!G(\textbf x^{(i)})\parallel _{2}] 
\end{split}
\label{optimisation with respect to generator}
\end{equation}
In our proposed segmentation network, the mapped generative net generates oil spill segmentation maps by drawing samples from the seminal distribution $\mathcal{P}_{sedi}(\textbf x)$, and the discriminative net discriminates the generated segmentation and the ground truth segmentation also by drawing samples. Therefore, in the training process, given a finite training set, the expectations in Eqs. (\ref{optimisation with respect to discriminator}) and (\ref{optimisation with respect to generator}) can be approximated using a minibatch of samples, and the loss functions of these two nets are given as follows:
\begin{equation}
\begin{split}
&\mathcal{L}_{L_{seg}}\!=\! \frac{1}{m} \sum_{i=1}^{m} \gamma_{seg}\big[\log D(\textbf y^{(i)})\!+\! \log\big(1\!\!-\!\!D(G(\textbf x^{(i)})))\big]
\end{split}
\label{loss function with respect to discriminator}
\end{equation}
and 
\begin{equation}
\begin{split}
\mathcal{L}_{L_{gen}}=&\frac{1}{m} \!\sum_{i=1}^{m}[\gamma_{seg}\log\big(1\!-\!D(G(\textbf x^{(i)})))\!+\gamma_{sreg}[\parallel \textbf y^{(i)}\\
&\!-\!G(\textbf x^{(i)})\parallel _{2}]\big] 
\end{split}
\label{loss function with respect to generator}
\end{equation}
where $m$ is the size of minibatch. Thus, the optimisation for the segmentation network is achieved by alternatively optimising the parameters of these two deep neural nets $G$ and $D$ with respect to their loss functions. Particularly, our proposed segmentation network performs accurate oil spill segmentation without requiring a large amount of oil spill SAR image training data. This beneficial from that we construct our proposed segmentation network with the collaboration of the seminal distribution representation of oil spill SAR images, by which the physical characteristics of the image data are modelled to instruct the mapped generative net producing accurate oil spill segmentations. Additionally, to further increase the similarity between the generated segmentation and the ground truth segmentation, a regularity term is formulated, which penalizes the differences between these two segmentations and thus contributes to achieve more accurate oil spill segmentation. Furthermore, the small amount of training data utilised provides a friendly way for automatic oil spill image segmentation, where the availability of oil spill SAR image data is limited in practice. The illustration of our proposed segmentation network for oil spill SAR image segmentation is described in Algorithm \ref{training for segmenting oil spill framework}.

\begin{algorithm}[htbp]
\caption{The Procedures of Training the Proposed Segmentation Method for Marine Oil Spill SAR Image Segmentation.}
\label{training for segmenting oil spill framework}
\KwIn{Marine oil spill image dataset including both the original images and the ground-truth segmentations.}
\KwOut{Oil spill segmentation maps.}
\begin{algorithmic}[1] 
    %\STATE Initialise parameters $\phi$ and $\theta$;
    \STATE \textbf {for} in the training iteration numbers \textbf {do}
    \STATE Optimisation for the maximisation step
    \STATE \quad $\bullet$ Draw mini-batch of $m$ examples from the prior\\ \qquad $\mathcal{P}_{sedi}(\textbf x)$;
    \STATE \quad $\bullet$ Draw minibatch of $m$ examples from the \\ \qquad ground truth distribution $\mathcal {G}_{gtdi}(\textbf y)$;
    \STATE \quad $\bullet$ Update the discriminator by ascending its \\ \qquad stochastic gradient:\\
     $\bigtriangledown _{\varphi_{ptd}}\frac{1}{m} \sum_{i=1}^{m} \gamma_{seg}\big[\log D(\textbf y^{(i)})\!+\! \log\big(1\!\!-\!\!D(G(\textbf x^{(i)})\!)\!)\!\big]$;
     \STATE Optimisation for the minimisation step
    \STATE \quad $\bullet$ Draw mini-batch of $m$ examples from the prior\\ \qquad $\mathcal{P}_{sedi}(\textbf x)$; 
    \STATE \quad $\bullet$ Draw minibatch of $m$ examples from the \\ \qquad ground truth distribution $\mathcal {G}_{gtdi}(\textbf y)$;
    \STATE \quad $\bullet$ Update the mapped generator by descending its\\
     \qquad stochastic gradient:\\
    \quad $\bigtriangledown _{\varphi_{ptg}}\frac{1}{m} \sum_{i=1}^{m}[\gamma_{seg}\log\big(1-D(G(\textbf x^{(i)})))$\\
    \quad \quad $+\gamma_{sreg}[\parallel \textbf y^{(i)}-G(\textbf x^{(i)})\parallel _{2}]\big]$;
    \STATE \textbf {end for};
%\RETURN Trained parameters $\bigtriangledown _{\varphi_{ptd}}$, $\bigtriangledown _{\varphi_{ptg}}$;
\end{algorithmic}
\end{algorithm}

\section{Experimental Work}
We have presented the details of constructing our proposed SRCNet in Sections \ref{the construction of oil spill segmentation network} and \ref{optimisation for segmentation}. To demonstrate the effectiveness of our proposed SRCNet for oil spill SAR image segmentation, in this section, we perform the experimental work by comparing its segmentation performance against several state-of-the-art segmentation methodologies. Particularly, to conduct a more comprehensive evaluation for our proposed SRCNet, the experimental dataset we utilise includes different types of oil spill SAR images, and the experimental evaluations from different metrics for the implemented segmentation techniques are given in the following sections.
\subsection{Experimental Preparations}
\label{description for image dataset}
\textit{\textbf{Image Dataset}}: In the experimental work, we conduct experimental evaluation for our proposed SRCNet using the image dataset that includes different types of oil spill SAR images. The SAR images utilised in our experimental work were captured by different sensors at different time, and the oil spills in the experimental images exhibit various areas with irregular shapes. Specifically, the experimental dataset includes SAR and ASAR images with VV, VH and HH polarization. In particular, the oil spill SAR images exploited in the experimental work are C-band oil spill SAR images from ERS-1 and ERS-2 satellites, C-band oil spill SAR images from radarsat satellite and C-band oil spill ASAR images from Envisat satellite, and we provide  specific descriptions for the experimental oil spill SAR images in Tables \ref{information of SAR sensors} and \ref{SAR image information of NOWPAP}, where the symbol "-" in Table \ref{SAR image information of NOWPAP} indicates unavailable information.

\begin{table}[htbp]
	\renewcommand\arraystretch{1.96}
	\centering
	\tabcolsep 0.11in
	\caption{SATELLITE SENSORS AND IMAGE DESCRIPTION.}
	\begin{tabular}{c|c|c|c}
		\hline
		\hline
		Satellite Sensor& Spatial Resolution & Waveband & Image Level  \\
		\hline	
		Radarsat-SAR & 50m x 50m  & C-band & 2 \\
		\hline
		ERS-1 SAR & 30m x 30m  & C-band & 2\\  
		\hline
		ERS-2 SAR & 30m x 30m  & C-band & 2\\
		\hline
		Envisat ASAR & 150m x 150m  & C-band & 2 \\
		%\hline
		%X-SAR & 6m x 6m  & X-band & 2 \\
		\hline
		\hline
	\end{tabular}
	\label{information of SAR sensors}	
\end{table}

\begin{table}[H]
	\renewcommand\arraystretch{1.97}
	\centering
	\tabcolsep 0.0023in
	\caption{DESCRIPTION OF OIL SPILL IMAGES.}
	\begin{tabular}{c|c|c|c}
		\hline
		\hline
		Capture Time & Satellite Sensor & Type of Oil Spills & Image Cover Ground \\
		\hline
		08.05.2010 12:01:25  & Radarsat & - & 207.4 km$^{2}$ \\
		\hline
		19.06.1995 02:30:55 & ERS-1 & - & 394 km$^{2}$ \\
		\hline
		19.06.1995 02:31:09 & ERS-1 & - & - \\
		%\hline
		%20.07.1997 02:14:41 & ERS-2 & Ship Oil Spill & -  \\ 
		\hline
		18.07.1998 02:03:24 & ERS-2 & Ship Oil Spill & - \\ 
		\hline
		17.03.2000 01:54:37& ERS-2 & Ship Oil Spill &- \\
		%29.05.2000 02:02:32 & ERS-2 & -&- \\
		%\hline
		%27.09.1999 02:02:05 & ERS-2 & - &- \\
		%\hline
		
		\hline
		26.07.1997 02:26:34 & ERS-2 & - &17.8x106m$^{2}$ \\
		\hline
		01.09.2008 01:11:51 & Envisat &  Ship Oil Spill &- \\
		\hline
% 		29.10.2008 13:43:33 & Envisat & - & - \\
% 		\hline
		15.08.2007 13:04:03 & Envisat & - &-  \\
		\hline
		28.07.2008 12:26:30  & Envisat & - & -\\
		%\hline
		%22.06.2010  & UAVSAR & Gulf Coast  \\
		\hline
		\hline
	\end{tabular}
	\label{SAR image information of NOWPAP}	
\end{table}

%  \subsubsection{Implementation Details} We perform the experimental work using the Tensorflow framework. Particularly, the segmentation model is trained on a NVIDIA Tesla P100 GPU with 16GB memory for 160 epochs with the batch size set as 1. We use the Adam optimizer with the learning rate set to $\alpha=1e-4$.

\subsection{Experimental Setup}
To demonstrate the effectiveness of our proposed SRCNet for segmenting oil spills in SAR images, we conduct experimental evaluations on different types of oil spill SAR images, and the specifications for image data information are given in Section \ref{description for image dataset}. The details of implementation training for oil spill SAR image segmentation are given as follows:
\subsubsection{Implementation Details} In the experimental work, we conduct the training for our proposed segmentation network with the input SAR images are scaled to the size of 128 $\times$ 128, and in the training dataset, the ground truth images are also provided. The ground truth images indicate the real oil spill areas clearly, and thus using the ground truth segmentations as criterion of training the segmentation netowrk helps to empower its capability in terms of correctly segmenting oil spills in SAR images.
\subsubsection{Training Operation} In the training of our proposed SRCNet for oil spill SAR image segmentation, we conduct the training work using the Tensorflow framework. Specifically, the training is conducted using Adam optimizer \cite{kingma2014adam} with the learning rate set to $\alpha=1e-4$, and the segmentation network is trained on a high performance computing sever with a NVIDIA Tesla P100 GPU with 16GB memory for 180 epochs, and the batch size is set as 1.
\subsubsection{Experimental Evaluation Metrics} In the experimental work, we conduct experimental evaluations from both qualitative and quantitative metrics. Following the works \cite{nieto2018two} \cite{garcia2017review} that the segmentation accuracy is a popular used evaluation criteria for image segmentation techniques. Thus, we evaluate the segmentation performance of the proposed segmentation network with respect to segmentation accuracy computed as follows: 

\begin{equation*}
\label{accuracy computation}
\text{Accuracy} = \frac{  \text{\# the number of correctly segmented pixels}}{ \text{\# the number of all pixels}}
\end{equation*}

Moreover, in the operation of image segmentation, the  Jaccard index (JCI) is regarded as an useful metric in evaluating the segmentation performance \cite{cantorna2019oil} \cite{muruganandham2016semantic}. Thus, to make further evaluation on our proposed method for the segmentation of SAR images, we conduct the segmentation evaluation in terms of JCI which is given as follows: 
\begin{equation*}
\text{JCI} = \frac{ \text{TP}}{ \text{TP + FP + FN}}
\end{equation*}
where TP, FP and FN represent the correctly segmented oil spill pixels, the background pixels that are incorrectly segmented as oil spills, and the oil spill pixels that are incorrectly segmented as background, respectively. Therefore, JCI is the intersection of the actual oil spill areas and the segmented areas, divided by the union of these two sections, and a more accurate oil spill segmentation has a higher computed JCI value.

\subsection{Comparison with Manual Assistant Segmentation Methods}
\label{comparison with manual assistant segmentation methods}
In this work, we propose an effective oil spill SAR image segmentation network named SRCNet which performs oil spill segmentation automatically. Thus, to evaluate the segmentation performance of our proposed SRCNet, in this part, we operate the experimental evaluation by comparing its segmentation against several representative manual assistant segmentation methodologies. These comparison methodologies include level set evolution (LSE) \cite{li2011level} strategy and the region active contour (RAC) model \cite{zhang2015level}. These two manual assistant segmentation methods operate oil spill image segmentation requiring a starting region that is manually devised to start the segmentation process, and thus the segmentation outcomes are correlated to the starting region to some extent. In this scenario, taking them as comparison provides a clear observation for the performance of our proposed segmentation technique. Particularly, to conduct a more detailed evaluation for the segmentation performance of our proposed method, we here utilise different types of oil spill SAR images as the experimental test dataset, and the segmentation results are shown in Fig. \ref{comparison with manual assistant segmentation methodologies}. In this figure, column (a) from top to bottom are the original ERS-1 SAR, ERS-2 SAR and the Envisat ASAR oil spill images, respectively, columns (b)-(d) are the corresponding segmentation results of the LSE, RAC and our proposed segmentation method, respectively, and column (e) shows the ground truth segmentations. Examining the segmentation results, it is clear that our proposed SRCNet achieves a more accurate oil spill SAR image segmentation.

Additionally, in image segmentation, the segmentation accuracy represents pixel-wise evaluation in terms of measuring the percentage of correctly segmented pixels \cite{cantorna2019oil}. Thus, to quantitatively evaluate the segmentation performance of our proposed method, we further compute its segmentation with respect to segmentation accuracy, and the computation results are shown in Table \ref{comparison with manual assistant segmentation methodologies}. In this Table, we can see that our proposed method achieves higher segmentation accuracy compared against the LSE and the RAC segmentation methods. This benefits from that our proposed SRCNet operates automatic oil spill image segmentation without requiring manually setting a starting region. This constructs a manual free manner for oil spill image segmentation and thus improves the segmentation efficiency. In addition, the manually setting results in the starting region is coarsely captured, which is not beneficial for accurate oil spill image segmentation because the oil spills in SAR images normally exhibit various areas with irregular shapes. Therefore, comparing our proposed SRCNet with the manual assistant methods, the experimental results validate that our proposed SRCNet achieves efficient and effective oil spill SAR image segmentation.

\begin{figure*}[t]
	\begin{center}
    %\raggedright
	\includegraphics[width=0.98\textwidth,height=0.41\textheight,center]{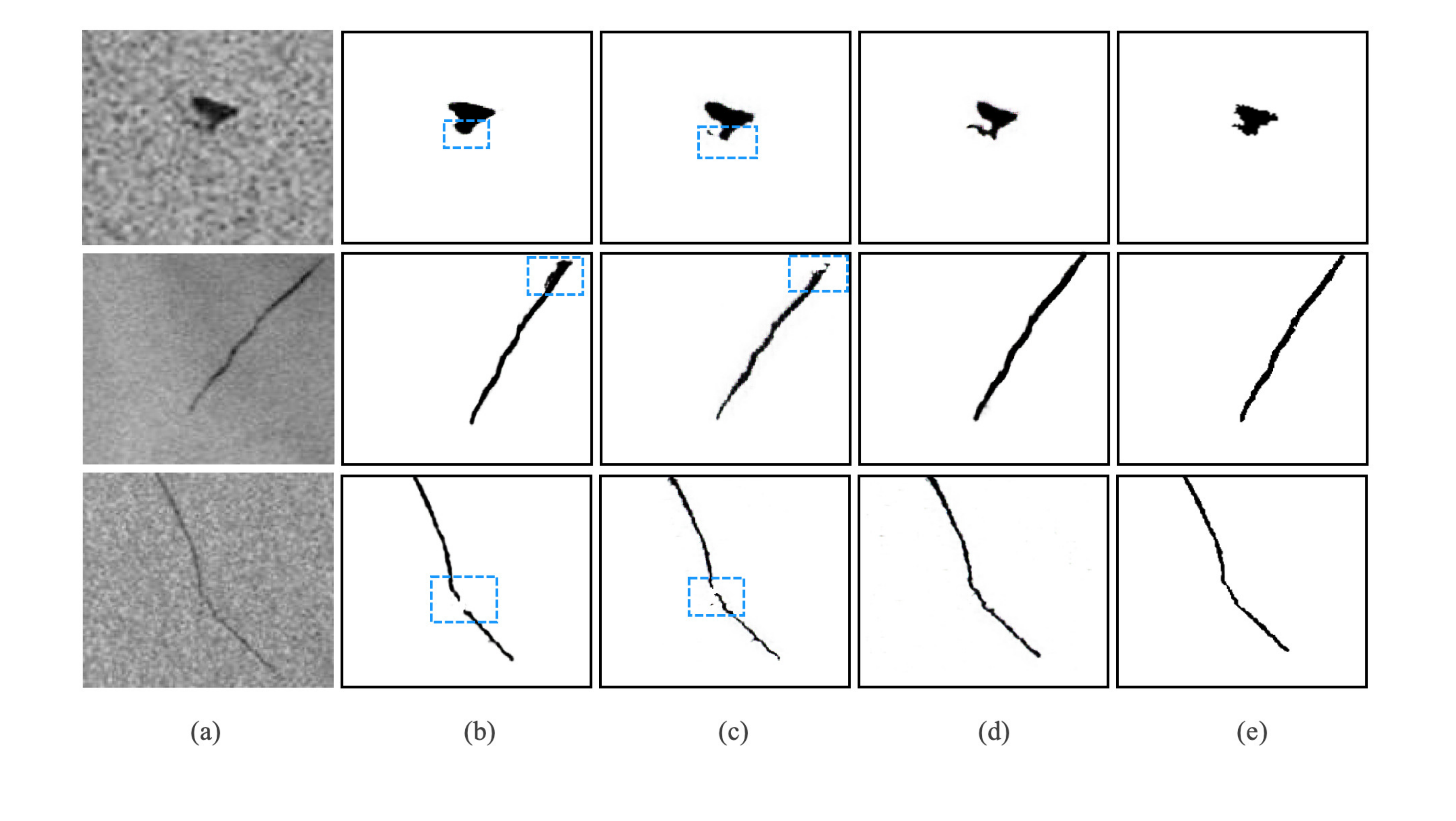}
	\end{center}
	\vspace*{-8.7mm}
	\caption{The segmentation of oil spill SAR images with the exploitation of different segmentation techniques. Specifically, (a) shows the original oil spill images, and from top to bottom are the ERS-1 SAR, ERS-2 SAR and Envisat ASAR oil spill images, respectively. (b)-(d) are the segmentation results of LSE, RAC and our proposed method, respectively. (e) shows the corresponding ground truth segmentations. The blue dashed-line boxes are utilised to indicate the incorrect segmentations.}
  \label{comparison with manual assistant segmentation methodologies}
\end{figure*}

\begin{table}[h]
 	\renewcommand\arraystretch{2.03}
 	\begin{center}
 	\tabcolsep 0.123in
 	\caption{THE SEGMENTATION ACCURACY OF LSE, RAC AND OUR PROPOSED METHOD.}
 	\begin{tabular}{c|ccc}
		\hline
 		\hline
 		\multirow{2}{*}{}{\diagbox[innerwidth=2.3cm]{Image}{Method}}
		&LSE & RAC & Our Method\\
		\hline
 		\hline
 		ERS-1 SAR & 0.8659 & 0.9031 & 0.9409 \\
 		\hline
 	    ERS-2 SAR & 0.9319 & 0.9308 & 0.9869 \\
 		\hline
 		Envisat ASAR & 0.9209 & 0.9386 & 0.9907 \\
 		\hline
 		\hline
 	\end{tabular}
 	\label{comparison with manual assistant segmentation methodologies}
 	\end{center}
 \end{table}

\subsection{Comparison with the Generative Adversarial Net}
\label{comparison with GAN}
\begin{figure}[htbp]
	\begin{center}
	%\vspace*{-28.6mm}
    %\raggedright
	\includegraphics[width=0.59\textwidth,height=0.47\textheight,center]{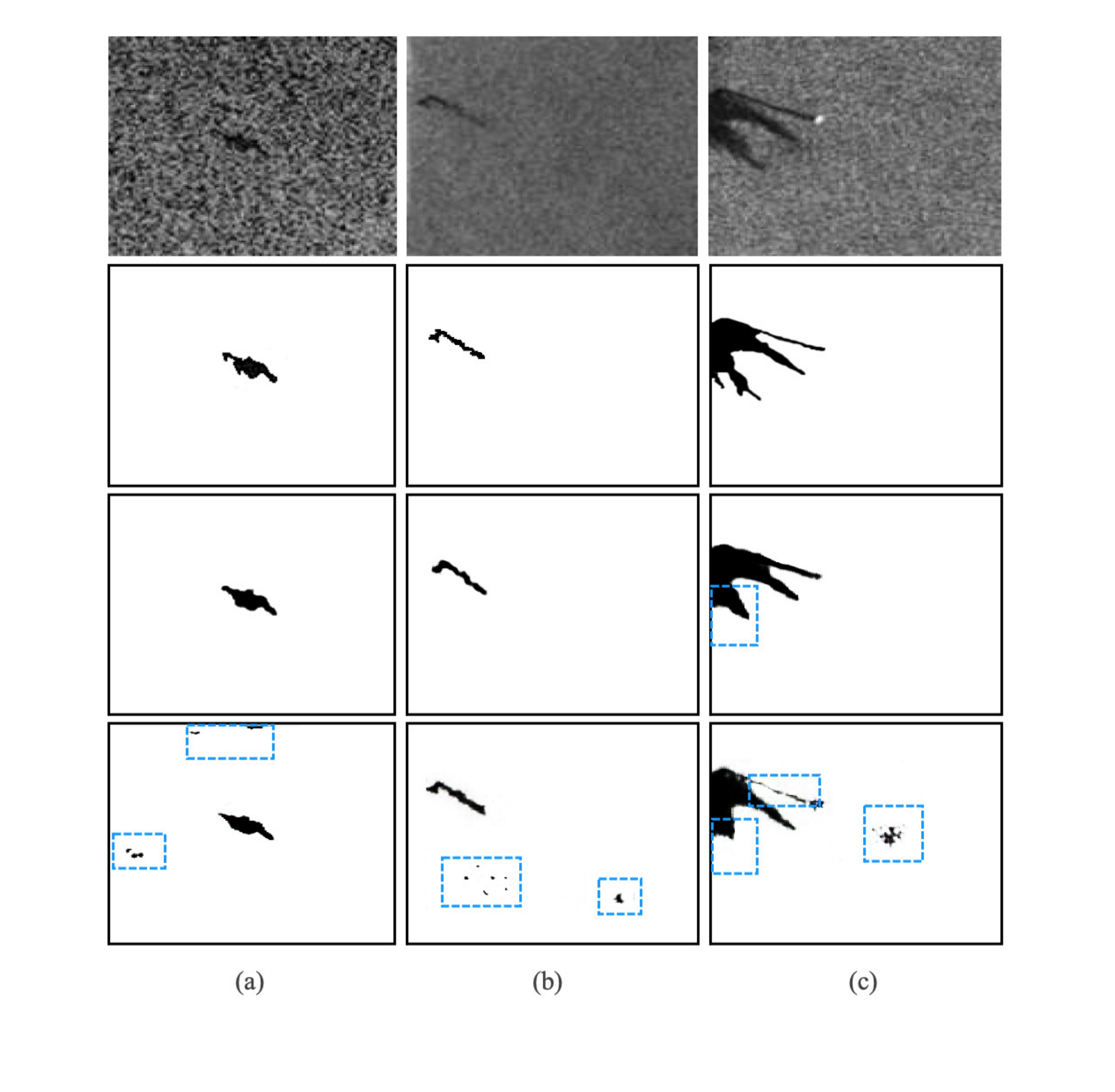}
	\end{center}
    \vspace*{-4.7mm}
	\caption{Comparison with the generative adversarial network. Specifically, in each column from top to bottom are the original oil spill SAR image, the ground truth segmentation and the segmentation results of our proposed SRCNet and GAN, respectively. The blue dashed line boxes are utilised to indicate the incorrect segmentations.}
  \label{comparison with deep neural technique}
\end{figure}

As described that in the application of deep neural networks for image processing, a striking point is the involvement of discriminative modules that play against the other modules so as for increasing the overall capability of the networks \cite{krizhevsky2017imagenet}. The generative adversarial net (GAN) \cite{goodfellow2020generative} is a representative deep neural network constructed with a discriminatve model which aims at estimating the probability of a sample that comes from real training data rather than the generated one. Therefore, to further evaluate the segmentation performance of our proposed method in terms of segmentation oil spills from SAR images, in this part, we operate the experimental work by comparing its segmentation performance with GAN. Specifically, to examine the  segmentation performance of GAN and our proposed SRCNet for different types of oil spill image segmentation, we here exploit ERS-1, ERS-2 and Radarsat oil spill SAR images as our experimental dataset, and the segmentation results are shown in Fig. \ref{comparison with deep neural technique}. In this figure, the first row from left to right are the ERS-1, ERS-2 and Radarsat oil spill SAR images, respectively, and in each column from top to bottom are the original oil spill SAR image, the ground truth segmentation, and the segmentations of our proposed method and GAN, respectively. Comparing the segmentation results of our proposed method and GAN with the ground truth segmentation, it is obvious that our proposed method achieves more accurate oil spill image segmentation. 

Additionally, to further evaluate the segmentation performance of our proposed method, we validate its segmentation with respect to segmentation accuracy shown in Table \ref{segmentation accuracy of deep neural techniques}. Examining the segmentation results in this table, it is clear that our proposed SRCNet achieves higher segmentation accuracy compared against GAN.

\begin{table}[htbp]
	\renewcommand\arraystretch{2.3}
	\begin{center}
	\tabcolsep 0.18in
	\caption{THE SEGMENTATION ACCURACY OF GAN AND OUR PROPOSED METHOD.}
	\begin{tabular}{c|ccc}
		\hline
		\hline
		\multirow{2}{*}{}{\diagbox[innerwidth=2.3cm]{Method}{Image}}
		&ERS-1 & ERS-2 & Radarsat\\
		\hline
		\hline
		Our Method & 0.9725 & 0.9910 & 0.9109 \\
		\hline
	    GAN & 0.8903 & 0.9271 & 0.7993 \\
		\hline
		\hline
	\end{tabular}
	\label{segmentation accuracy of deep neural techniques}
	\end{center}
	\vspace*{-4.7mm}
\end{table}

\subsection{Comprehensive Evaluation for Segmentation Different Types of Oil Spill Images}
\begin{figure*}[htbp]
	\begin{center}
	%\vspace*{-46.6mm} 
    %\raggedright
	\includegraphics[width=1.23\textwidth,height=0.94\textheight,center]{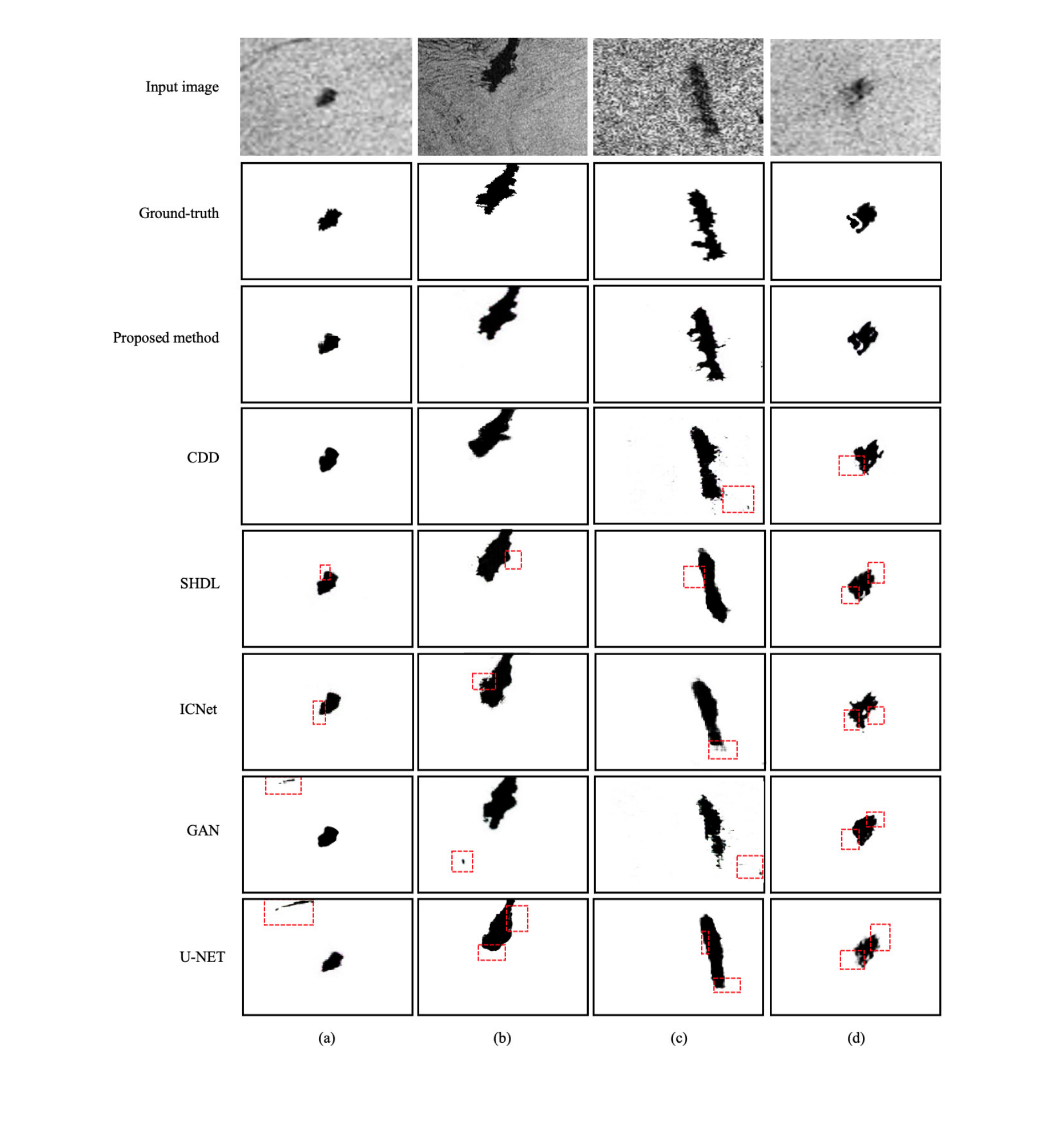}
	\end{center}
	\vspace*{-16.0mm}
	\caption{Segmentation of ERS-1 oil spill SAR images with the exploitation of different segmentation methodologies. Specifically, columns (a)-(d) show oil spill SAR images and their corresponding segmentation results with the implemented segmentation techniques. Particularly, from top to bottom in each column are the original oil spill SAR image, the ground-truth segmentation, the segmentation with our proposed method, CDD, SHDL, ICNet, GAN and the U-NET segmentation methodologies respectively. The red boxes are utilised to indicate the incorrect segmentations.}
  \label{the segmentation of ERS-1 oil spill SAR images}
\end{figure*}

\begin{figure*}[htbp]
	\begin{center}
	%\vspace*{-46.6mm} 
    %\raggedright
	\includegraphics[width=1.23\textwidth,height=0.94\textheight,center]{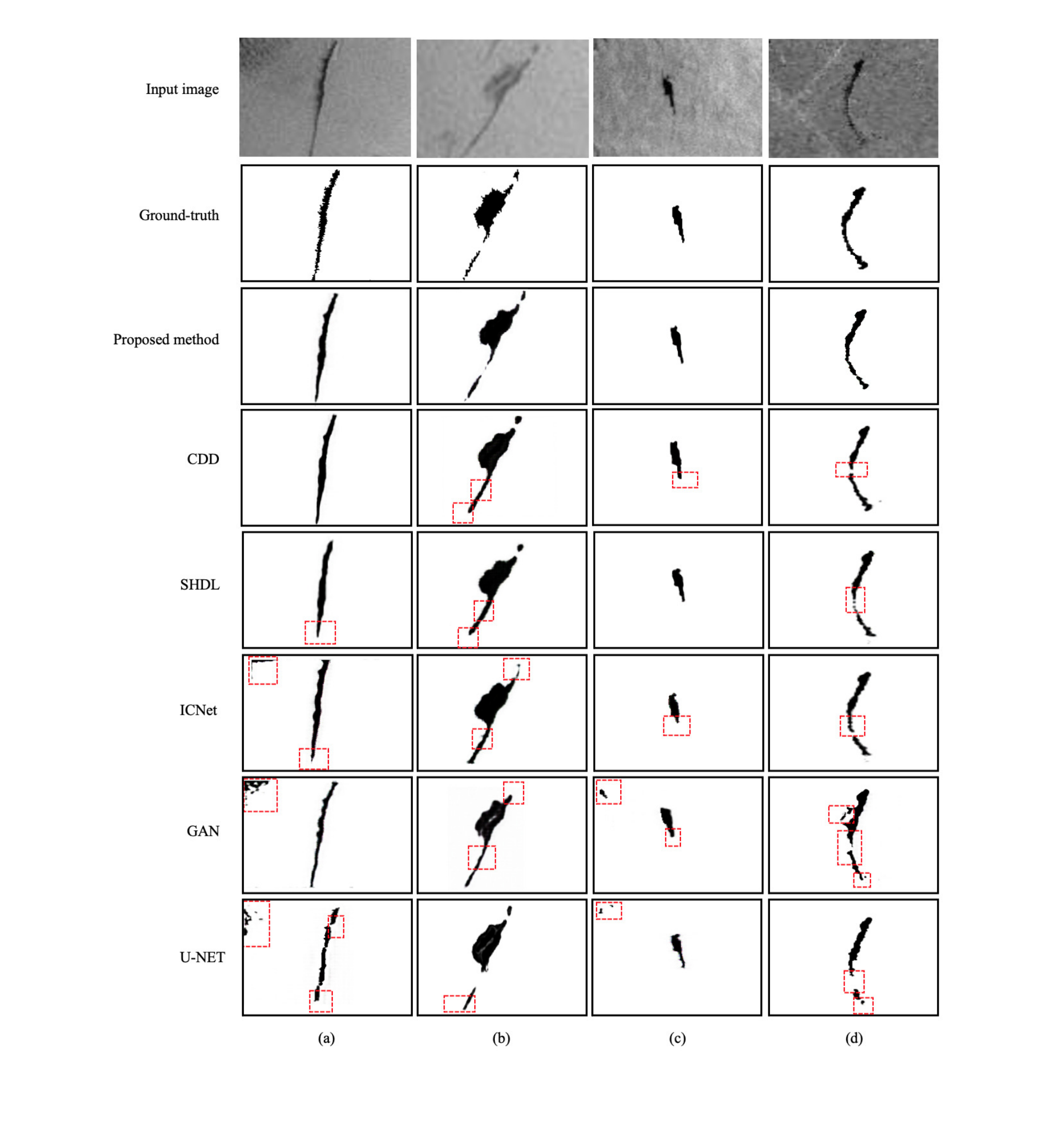}
	\end{center}
	\vspace*{-16.0mm}
	\caption{Segmentation of ERS-2 oil spill SAR images with the exploitation of different segmentation methodologies. Specifically, columns (a)-(d) show oil spill SAR images and their corresponding segmentation results with the implemented segmentation techniques. Particularly, from top to bottom in each column are the original oil spill SAR image, the ground-truth segmentation, the segmentation with our proposed method, CDD, SHDL, ICNet, GAN and the U-NET segmentation methodologies respectively. The red boxes are utilised to indicate the incorrect segmentations.}
  \label{the segmentation of ERS_2 oil spill SAR images}
\end{figure*}

\begin{figure*}[htbp]
	\begin{center}
	%\vspace*{-46.6mm}
    %\raggedright
	\includegraphics[width=1.23\textwidth,height=0.94\textheight,center]{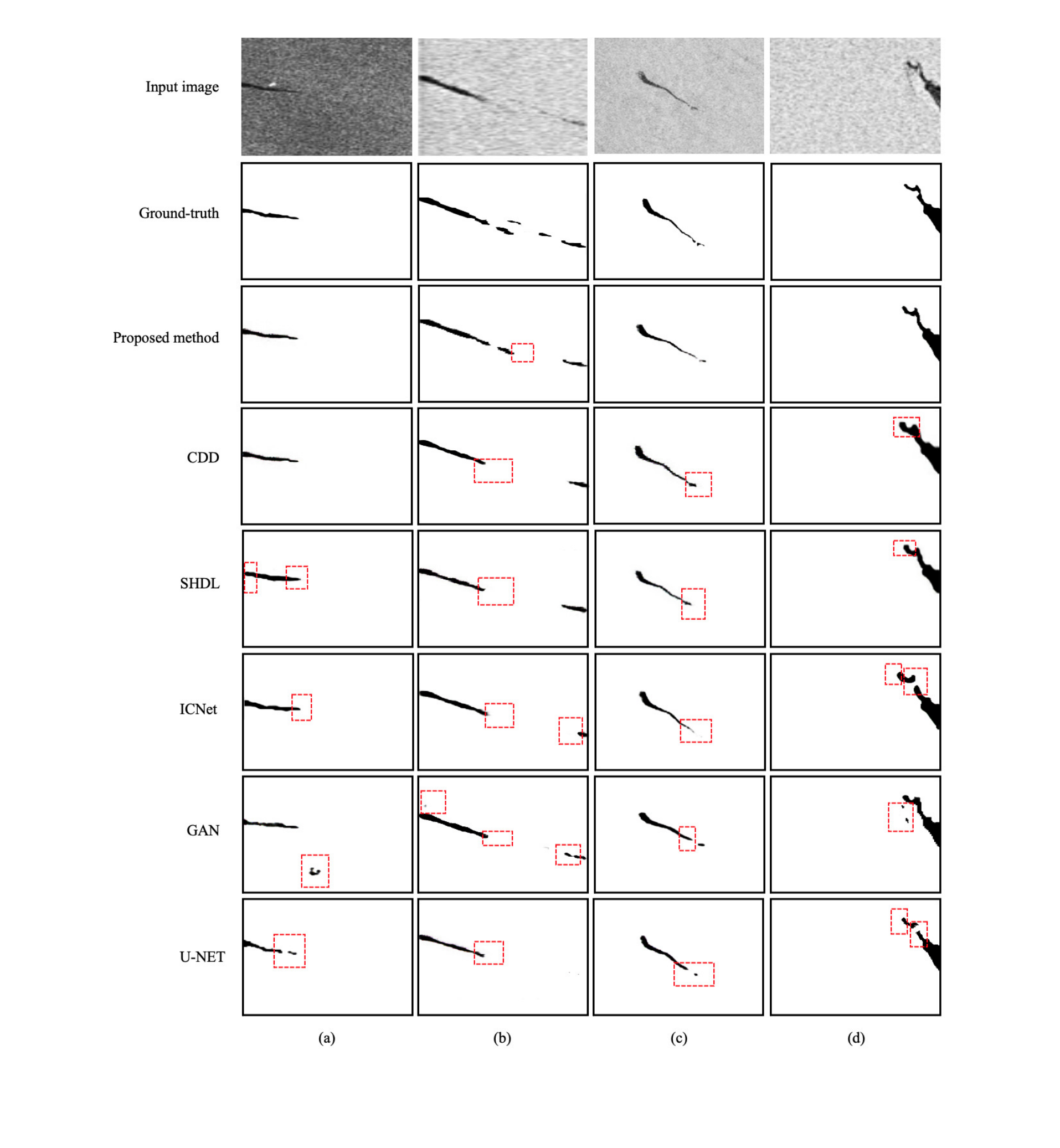}
	\end{center}
	\vspace*{-16.0mm}
	\caption{Segmentation of Radarsat oil spill SAR images with the exploitation of different segmentation methodologies. Specifically, columns (a)-(d) show oil spill SAR images and their corresponding segmentation results with the implemented segmentation techniques. Particularly, from top to bottom in each column are the original oil spill SAR image, the ground-truth segmentation, the segmentation with our proposed method, CDD, SHDL, ICNet, GAN and the U-NET segmentation methodologies respectively. The red boxes are utilised to indicate the incorrect segmentations.}
  \label{the segmentation of radasat oil spill SAR images}
\end{figure*}

We have initially evaluate the segmentation performance of our proposed SRCNet by comparing its segmentation against manual assistant methods and the generative adversarial network in Sections \ref{comparison with manual assistant segmentation methods} and \ref{comparison with GAN}, separately, and the experimental evaluations in these two parts demonstrated the effectiveness of our proposed segmentation network. Besides, to evaluate the segmentation performance of our proposed method one step further, in this part, we conduct a more comprehensive evaluation in terms of segmenting oil spill images with more evaluation metrics. The comprehensive evaluations enable the examination for our proposed method in a more wide scope and therefore well validate its segmentation performance. Specifically, we here conduct the experimental evaluation by comparing its segmentation against several representative image segmentation methods using ERS-1 SAR, ERS-2 SAR and Radarsat oil spill images as the experimental dataset, and the segmentation results are shown in Figs. \ref{the segmentation of ERS-1 oil spill SAR images} - \ref{the segmentation of radasat oil spill SAR images}. In details, the comparison methodologies include capacitory discrimination distance (CDD) learning technique \cite{yu2018oil}, squared helinger divergence learning (SHDL) \cite{yu2018oil}, image cascade network  (ICNet) for image segmentation \cite{zhao2018icnet}, GAN and the U-Net segmentation neural network \cite{ronneberger2015u}. Specifically, Figs. \ref{the segmentation of ERS-1 oil spill SAR images}, \ref{the segmentation of ERS_2 oil spill SAR images} and \ref{the segmentation of radasat oil spill SAR images} show the segmentation for ERS-1, ERS-2 and Radarsat oil spill SAR images respectively using the comparison methodologies and our proposed SRCNet. Particularly, in each figure, from top to bottom are the original oil spill SAR images, the ground truth segmentation, and the segmentation results of our proposed method, CDD, SHDL, ICNet, GAN and the U-Net, separately. Examining these segmentation results, it is obvious that our proposed SRCNet performs more accurate oil spill SAR image segmentation compared against the other employed segmentation methods.

Besides the qualitative segmentation evaluation, to further validate the segmentation performance of our proposed SRCNet and the other employed segmentation methodologies, we here conduct the quantitative experimental evaluation in terms of the segmentation accuracy shown in Tables \ref{ERS-1 oil spill SAR image segmentation accuracy}, \ref{ERS-2 oil spill SAR image segmentation accuracy} and \ref{Radarsat oil spill SAR image segmentation accuracy}. Specifically, Tables \ref{ERS-1 oil spill SAR image segmentation accuracy}, \ref{ERS-2 oil spill SAR image segmentation accuracy} and \ref{Radarsat oil spill SAR image segmentation accuracy} show the segmentation accuracy of the implemented segmentation techniques (including our proposed SRCNet) for ERS-1, ERS-2 and Radarsat oil spill SAR image segmentation, respectively. Examining the quantitative experimental results in these Tables, it is clear that our proposed segmentation method achieves higher segmentation accuracy than the other employed segmentation techniques. 

\begin{table*}[htbp]
	\renewcommand\arraystretch{2.3}
	\begin{center}
	\tabcolsep 0.18in
	\caption{SEGMENTATION ACCURACY OF ERS-1 OIL SPILL SAR IMAGES.}
	\begin{tabular}{c|cccccc}
		\hline
		\hline
		\multirow{2}{*}{}{\diagbox[innerwidth=2.6cm]{Image}{Method}}
		&U-Net & GAN & ICNet & SHDL & CDD & SRCNet (Ours)\\
		\hline
		\hline
		(a) & 0.8958 & 0.9307 & 0.9682 & 0.9703 & 0.9728 & \textbf{0.9794} \\
		\hline
		(b) & 0.9026 & 0.9458 & 0.9473 & 0.9582 & 0.9831 & \textbf{0.9869} \\
		\hline
		(c) & 0.9021 & 0.8997 & 0.9228 & 0.9231 & 0.9308 & \textbf{0.9847} \\
		\hline
		(d) & 0.8923 & 0.9037 & 0.9249 & 0.9204 & 0.9385 &  \textbf{0.9731} \\
		\hline
		\hline
	\end{tabular}
	\label{ERS-1 oil spill SAR image segmentation accuracy}
	\end{center}
\end{table*}

\begin{table*}[htbp]
	\renewcommand\arraystretch{2.3}
	\begin{center}
	\tabcolsep 0.18in
	\caption{SEGMENTATION ACCURACY OF ERS-2 OIL SPILL SAR IMAGES.}
	\begin{tabular}{c|cccccc}
		\hline
		\hline
		\multirow{2}{*}{}{\diagbox[innerwidth=2.6cm]{Image}{Method}}
		&U-Net & GAN & ICNet & SHDL & CDD & SRCNet (Ours)\\
		\hline
		\hline
		(a) & 0.8959 & 0.8981 & 0.9254 & 0.9683 & 0.9721 & \textbf{0.9883} \\
		\hline
		(b) & 0.9268 & 0.9145 & 0.9196 & 0.9239 & 0.9247 & \textbf{0.9893} \\
		\hline
		(c) & 0.9521 & 0.9430 & 0.9731 & 0.9828 & 0.9821 & \textbf{0.9908} \\
		\hline
		(d) & 0.9233 & 0.9286 & 0.9679 & 0.9683 & 0.9668 &  \textbf{0.9901} \\
		\hline
		\hline
	\end{tabular}
	\label{ERS-2 oil spill SAR image segmentation accuracy}
	\end{center}
\end{table*}

\begin{table*}[htbp]
	\renewcommand\arraystretch{2.3}
	\begin{center}
	\tabcolsep 0.18in
	\caption{SEGMENTATION ACCURACY OF RADARSAT OIL SPILL SAR IMAGES.}
	\begin{tabular}{c|cccccc}
		\hline
		\hline
		\multirow{2}{*}{}{\diagbox[innerwidth=2.6cm]{Image}{Method}}
		&U-Net & GAN & ICNet & SHDL & CDD & SRCNet (Ours)\\
		\hline
		\hline
		(a) & 0.9685 & 0.9507 & 0.9723 & 0.9782 & 0.9886 & \textbf{0.9923} \\
		\hline
		(b) & 0.8731 & 0.9054 & 0.9003 & 0.9238 & 0.9221 & \textbf{0.9647} \\
		\hline
		(c) & 0.9623 & 0.9689 & 0.9726 & 0.9808 & 0.9821 & \textbf{0.9897} \\
		\hline
		(d) & 0.9543 & 0.9657 & 0.9639 & 0.9743 & 0.9725 &  \textbf{0.9931} \\
		\hline
		\hline
	\end{tabular}
	\label{Radarsat oil spill SAR image segmentation accuracy}
	\end{center}
\end{table*}

\begin{figure*}[htbp]
  \begin{center}
  %\vspace*{-44mm}
  %
  \subfigure[]{
  \includegraphics[width=0.31\textwidth,height=0.17\textheight]{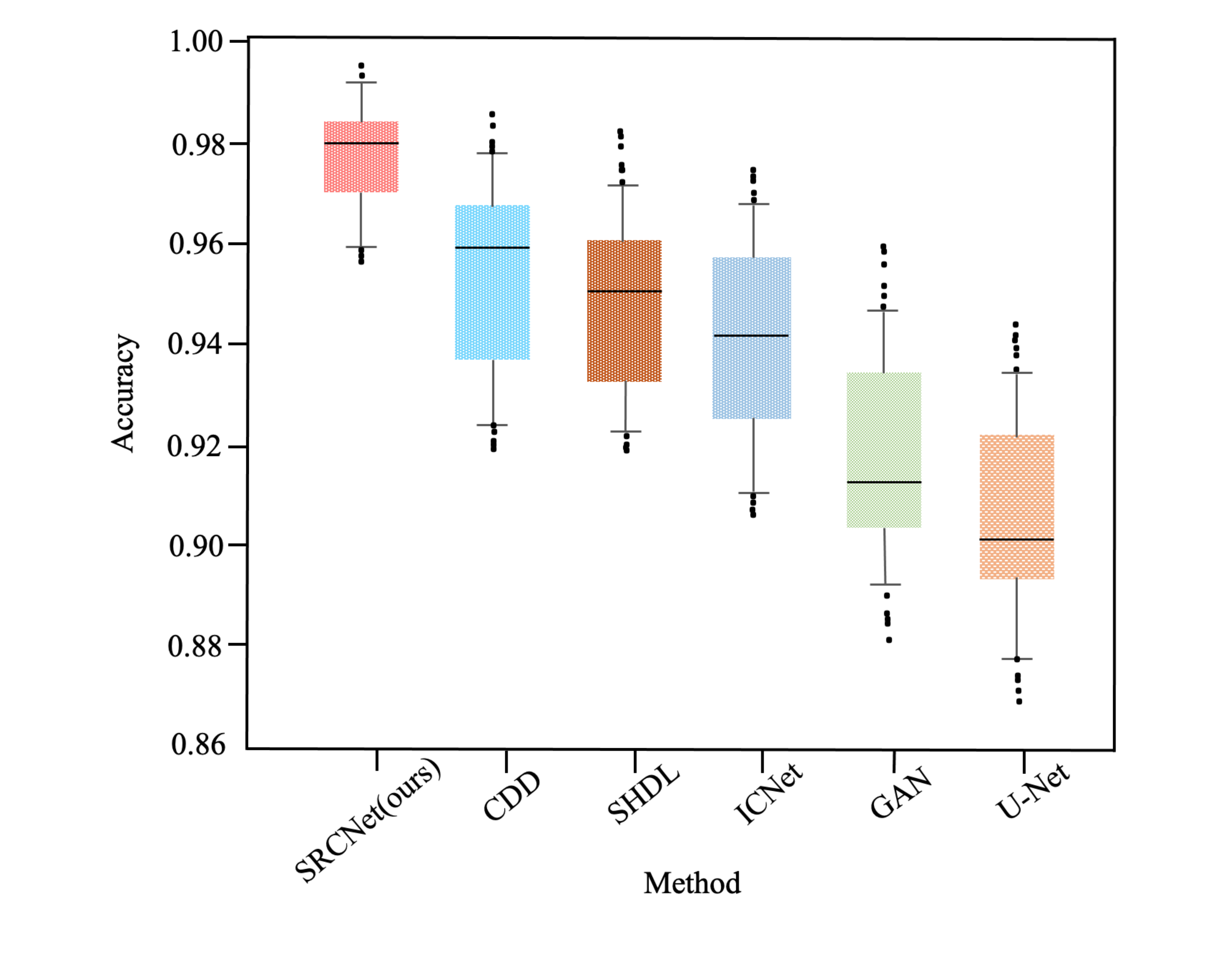}}
  \subfigure[]{
  \includegraphics[width=0.31\textwidth,height=0.17\textheight]{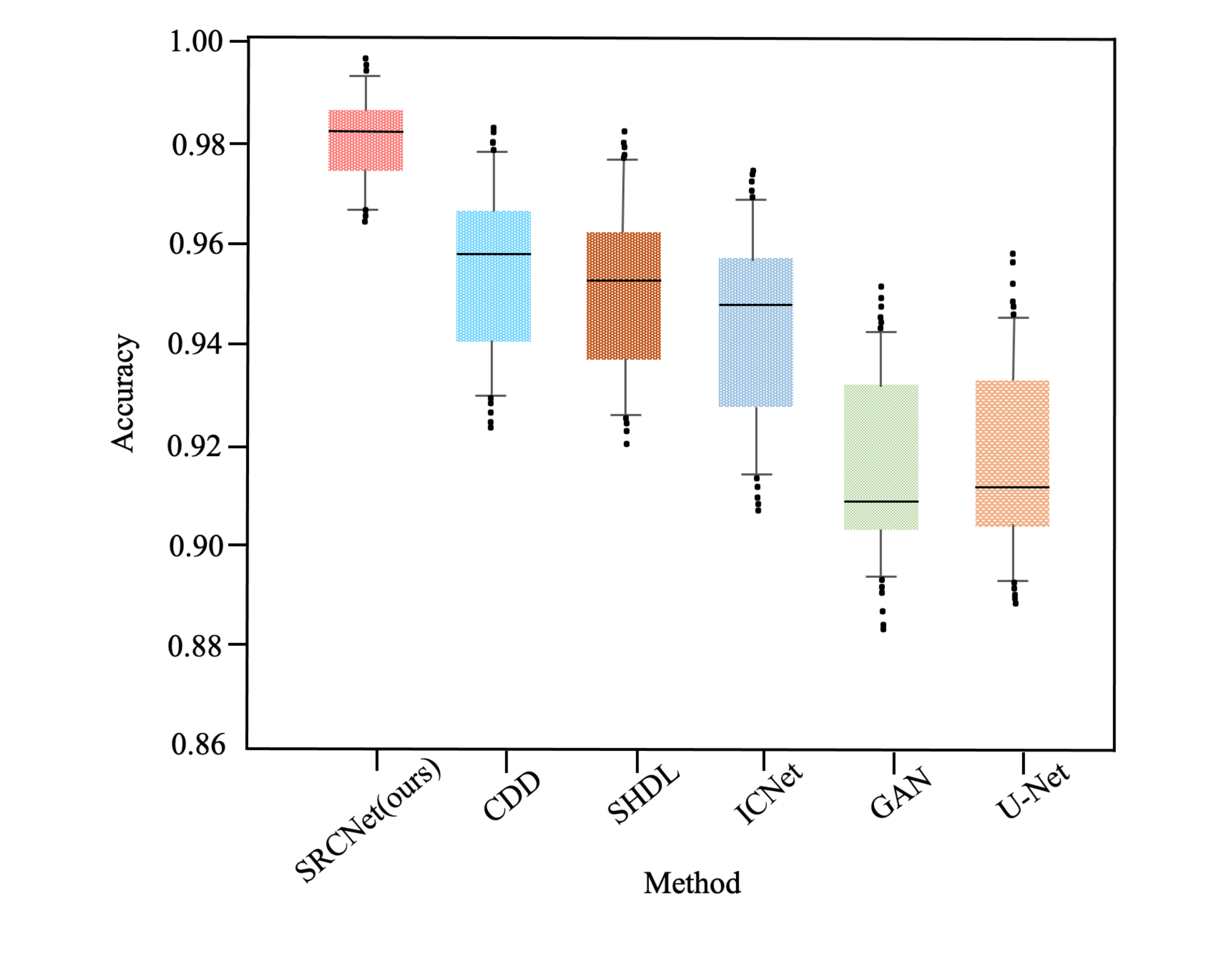}}
  \subfigure[]{
  \includegraphics[width=0.31\textwidth,height=0.17\textheight]{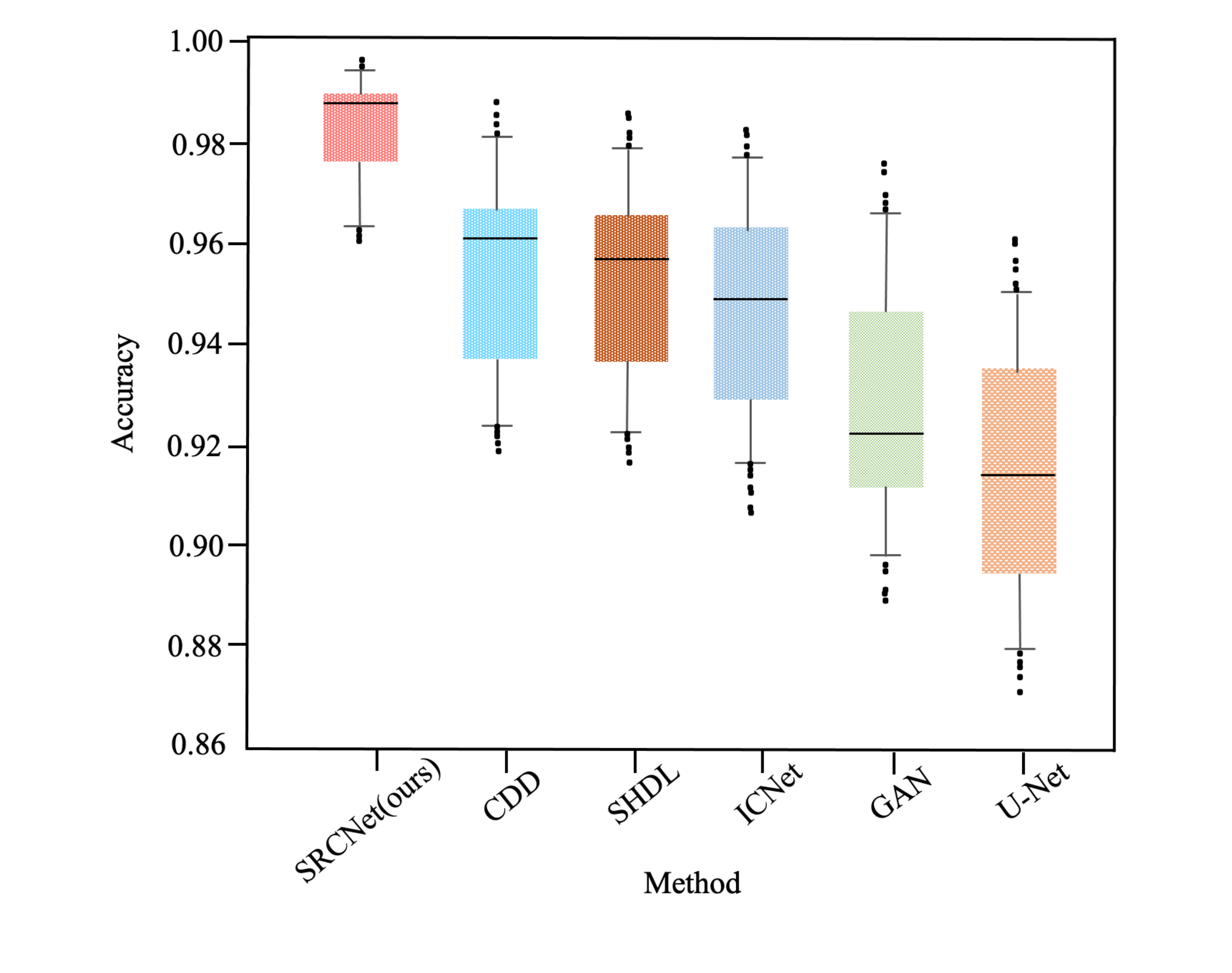}}
  \end{center}
  \vspace*{-2mm}
  \caption{Segmentation accuracy distributions with respect to different types of oil spill images. Specifically, (a), (b) and (c) illustrate the distributions of the segmentation accuracy that corresponds to each method for ERS-1, ERS-2 and Radarsat oil spill image segmentation, respectively. The rectangles corresponding to the strategies depict the accuracy values distributed between the minimum and maximum values (these two values are shown with bars on bottom-up sides of the rectangles), and the black dots are the segmentation outliers that fall out the interval of the two bars.}
  \label{box plot representation} 
\end{figure*}

The qualitative segmentation results shown in Figs. \ref{the segmentation of ERS-1 oil spill SAR images}, \ref{the segmentation of ERS_2 oil spill SAR images} and \ref{the segmentation of radasat oil spill SAR images} and the computation segmentation accuracy shown in Tables \ref{ERS-1 oil spill SAR image segmentation accuracy}, \ref{ERS-2 oil spill SAR image segmentation accuracy} and \ref{Radarsat oil spill SAR image segmentation accuracy} demonstrate the effectiveness of our proposed segmentation method for different types of oil spill SAR image segmentation. Together with these segmentation evaluations, to further validate the segmentation performance of our proposed SRCNet for oil spill SAR image segmentation, we compute the segmentation results with respect to segmentation accuracy statistically. The statistical evaluation gives an additional quantitative way for describing the segmentation performance and the stability of the employed methods, and the computation results are illustrated with box plots showing in Fig. \ref{box plot representation}. Specifically, Fig. \ref{box plot representation}(a), (b) and (c) show the statistical demonstration of the computed accuracy distribution ranges of the implemented segmentation methods for ERS-1, ERS-2 and the Radarsat oil spill SAR image segmentation, respectively. In particular, the coloured boxes in each sub-figure are utilised to depict the clustered accuracy values, where the bars on the bottom-up sides of the boxes indicate the levels of the minimum and maximum values, and the black dots are the segmentation outliers that fall out the interval of the two bars. Examining the statistical computation results, it is obvious that our proposed SRCNet achieves higher accuracy of oil spill SAR image segmentation with a smaller number of outliers. This demonstrates that our proposed SRCNet achieves more accurate and stable oil spill image segmentation compared against the other implemented segmentation methodologies. 

\begin{figure}[h]
	\begin{center}
	%\vspace*{-28.6mm}
    %\raggedright
	\includegraphics[width=0.47\textwidth,height=0.26\textheight,center]{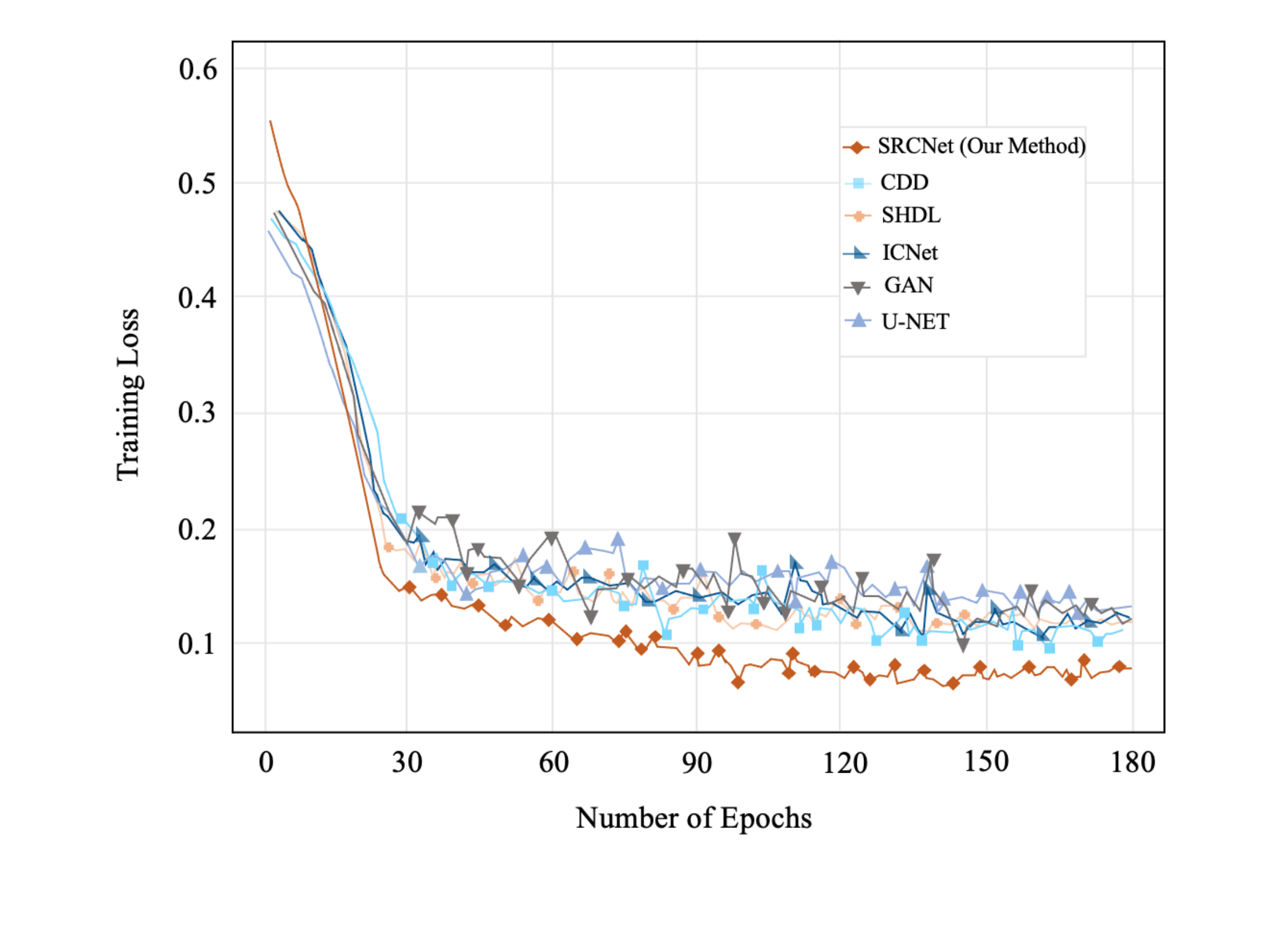}
	\end{center}
    \vspace*{-4.7mm}
	\caption{The learning curve of our proposed SRCNet and the other implemented segmentation networks.}
  \label{learning curve of the deep neural techniques}
\end{figure}

Our proposed SRCNet operates oil spill SAR image segmentation with the collaboration of the seminal distribution representation that describes oil spill SAR images. The seminal representation originates from SAR imagery for observing oceanic scenes and thus it provides pixel-wise physical information of oil spill SAR images. Therefore, in the training for oil spill SAR image segmentation, the collaborated seminal representation encourages efficient learning for effective oil spill segmentation. In this scenario, to demonstrate the training efficiency of our proposed SRCNet for operating oil spill SAR image segmentation, we illustrate the learning curve of our proposed SRCNet and the other implemented segmentation networks shown in Fig. \ref{learning curve of the deep neural techniques}. In this figure, it is clear that our proposed SRCNet converges to a lower training loss with less epochs compared against the other implemented segmentation networks, and with the training epoch increasing, our proposed segmentation network fine tunes the segmentation map and thus achieves accurate oil spill image segmentation. 

\begin{figure*}[htbp]
  \begin{center}
  %\vspace*{-44mm} 
  %
  \subfigure[]{
  \includegraphics[width=0.31\textwidth,height=0.16\textheight]{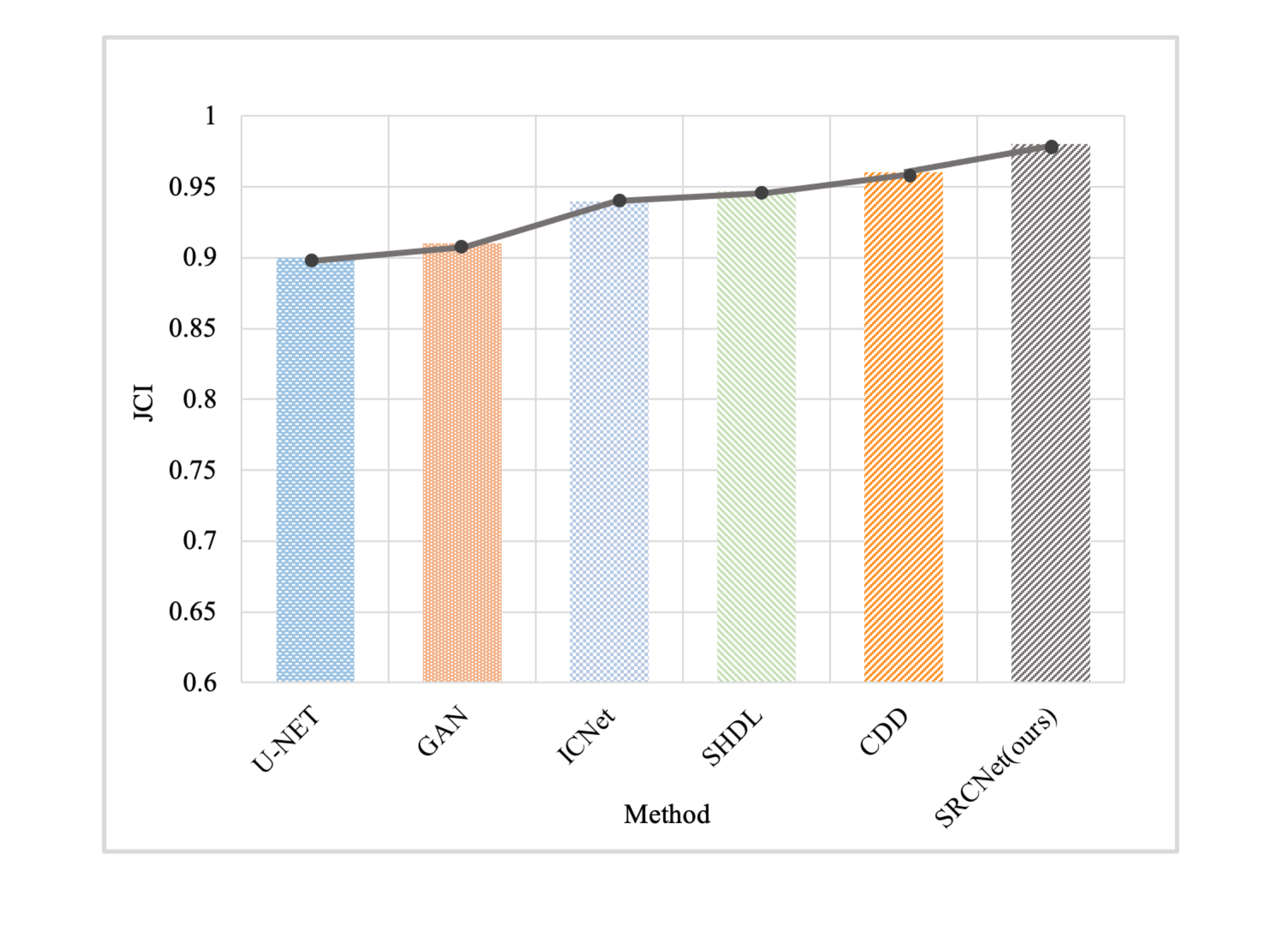}}
  \subfigure[]{
  \includegraphics[width=0.31\textwidth,height=0.16\textheight]{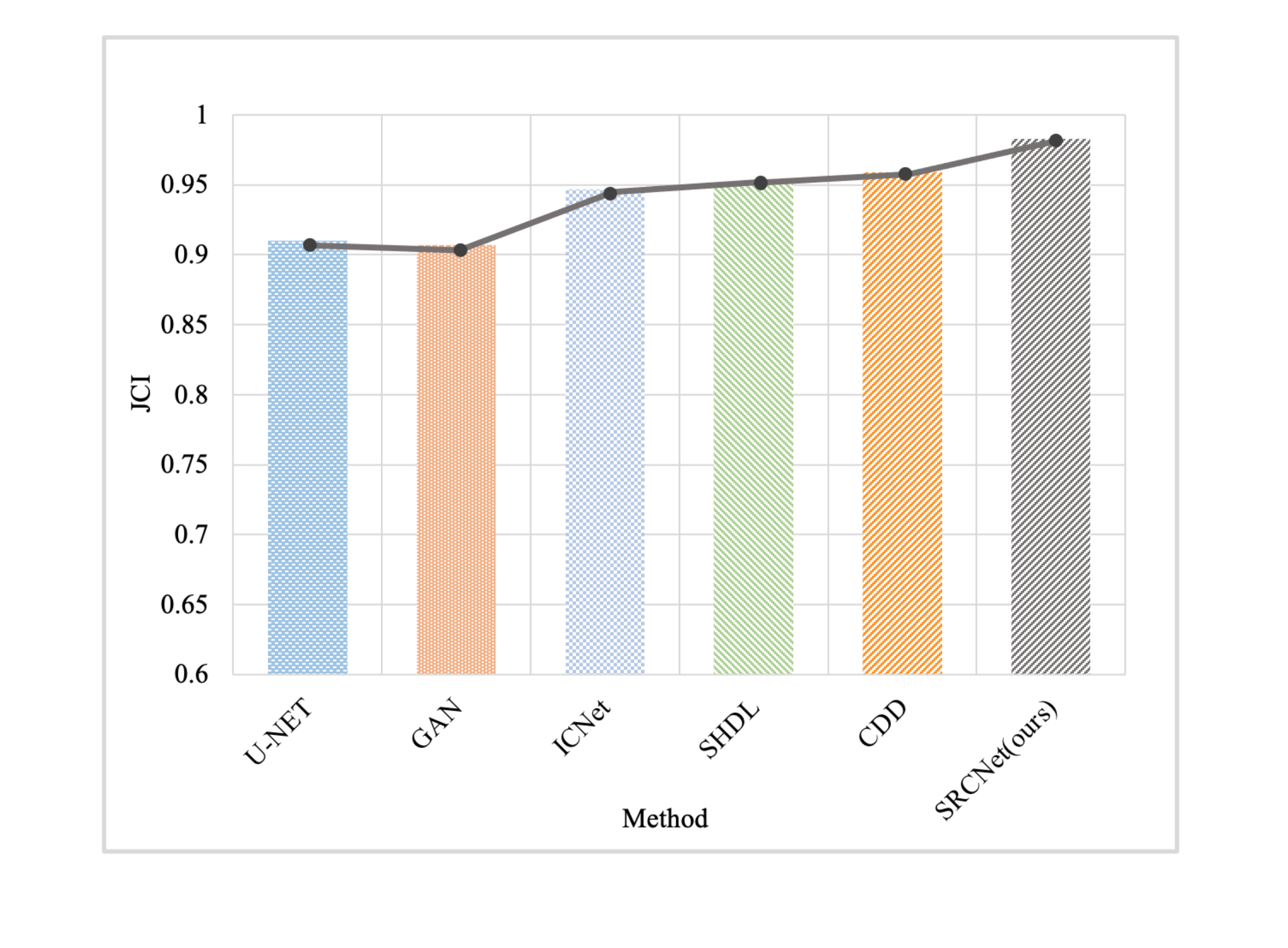}}
  \subfigure[]{
  \includegraphics[width=0.31\textwidth,height=0.16\textheight]{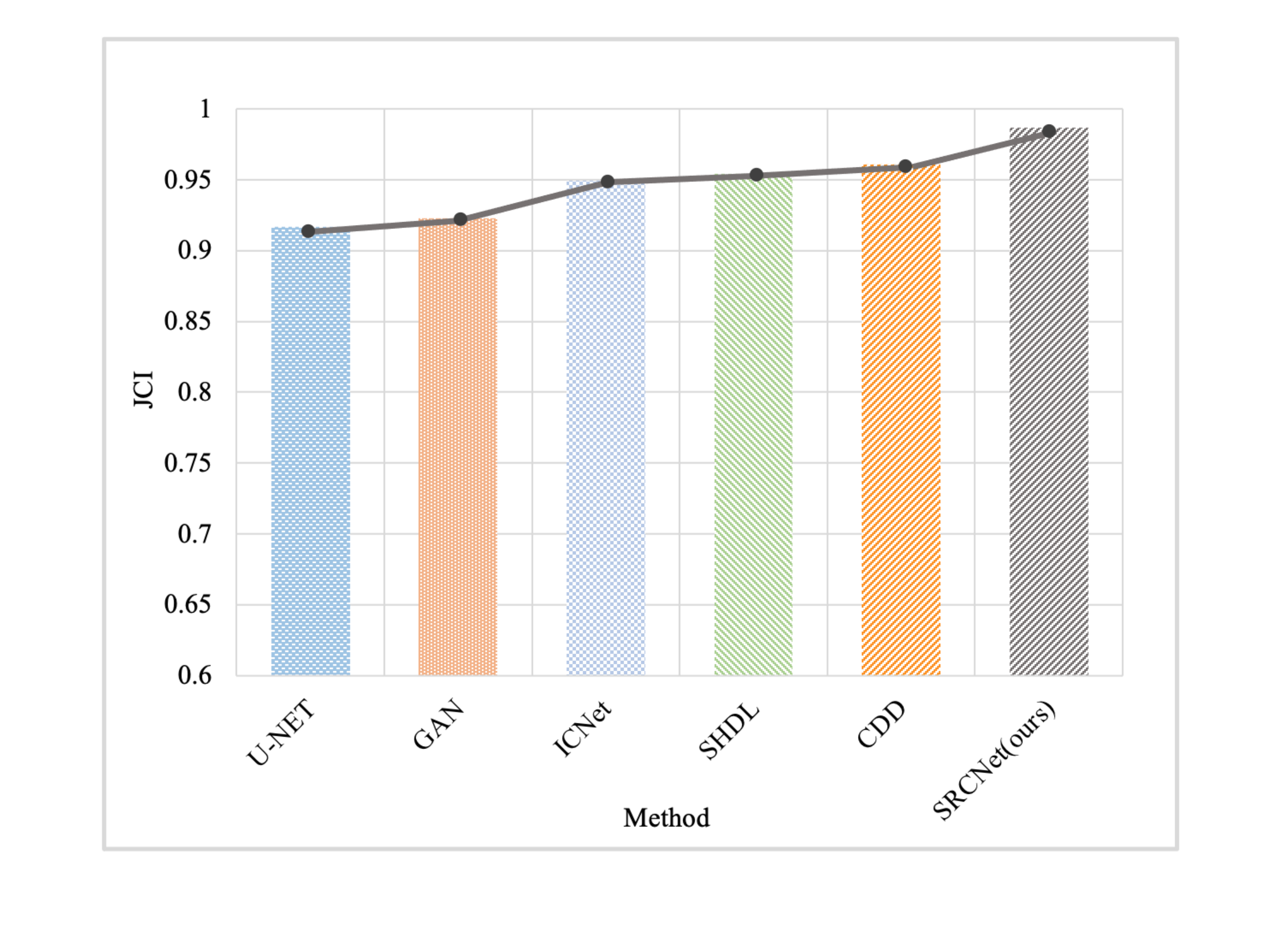}}
  \end{center}
\vspace*{-2mm}
  \caption{The segmentation performance of the exploited segmentation networks for oil spill SAR image segmentation with respect to Jaccard index (JCI). In this figure, from (a) to (c) indicate the JCI of the employed methods for ERS-1, ERS-2 and Radarsat oil spill SAR image segmentation, respectively, and the columns in each sub-figure indicate the level of computed JCI of the exploited segmentation networks. Particularly, the dark points on each column are utilised to mark the levels of the computed JCI of the employed segmentation methods, and the lines that connect the dark points show the trend of the segmentation performance of the alternative methods.}
  \label{Jaccard index computation} 
\end{figure*}

Additionally, in the implementation of oil spill SAR image segmentation, an accurate segmentation is that the segmented oil spill areas tightly approach the ground truth segmentation areas. The closeness of these two areas is mathematically described as the intersection. Besides, in the operation of deep neural networks for image segmentation, the Jaccard index (JCI) is a key metric that indicates the segmentation performance of the employed segmentation networks \cite{cantorna2019oil} \cite{long2015fully}. In this scenario, to further evaluate the segmentation performance of the implemented segmentation networks (including our proposed SRCNet), we demonstrate the computed JCI of each segmentation network for oil spill SAR image segmentation shown in Fig. \ref{Jaccard index computation}. Specifically, Fig. \ref{Jaccard index computation}(a), (b) and (c) illustrate the computation results of JCI for ERS-1, ERS-2 and Radarsat oil spill SAR image segmentation with the exploitation of the alternative segmentation networks, respectively, and the columns in each sub-figure indicate the levels of the computed JCI. From the description and the computation representation of JCI that a higher level result represents a more accurate oil spill image segmentation. For clarity, we mark the levels of JCI with dark points and utilise lines to connect these point to show the trend of the segmentation performance of the alternative methods. Examining the columns in each sub-figure, it is clear that our proposed SRCNet achieves a higher level of JCI compared against the other segmentation networks. This demonstrates that the segmented oil spill areas from our proposed SRCNet tightly approach the ground truth segmentation areas, and therefore performs accurate oil spill segmentation.

\section{Conclusion}
We have presented the procedures of constructing our proposed SRCNet marine oil spill SAR image segmentation network, which performs effective oil spill image segmentation in an automatic way. Specifically, in our proposed segmentation network, two deep neural nets are structured that work in a competition manner, with one is a mapped generative net which aims at producing accurate oil spill segmentation maps by drawing examples from the seminal distribution of oil spill SAR images, and the other is the discriminative net which strives to distinguish the generated segmentation and the ground truth segmentation. In this scenario, in the training process, the segmentation network convergence with the generated segmentation is as accurate as the ground truth segmentation and therefore fooling the discriminative net  unable to distinguish between the generated and the ground truth segmentations. Thus, the efficiency and effectiveness of the segmentation network is closely corrected to the generative net, and to render effective and efficient oil spill segmentation, we explore the seminal representation of oil spill SAR images, and incorporate it into the segmentation model. The seminal representation originates from SAR imagery, describing the internal characteristics of oil spill SAR images. Therefore, feeding the mapped generative net with examples drawing from the seminal representation enabling it to produce accurate oil spill segmentation maps in high efficiency. This establishes effective and efficient oil spill segmentation of our proposed segmentation network. Experimental evaluations from different metrics validate that our proposed segmentation network performs accurate oil spill segmentations for different types of oil spill SAR images.

% if have a single appendix:
%\appendix[Proof of the Zonklar Equations]
% or
%\appendix  % for no appendix heading
% do not use \section anymore after \appendix, only \section*
% is possibly needed

% use appendices with more than one appendix
% then use \section to start each appendix
% you must declare a \section before using any
% \subsection or using \label (\appendices by itself
% starts a section numbered zero.)
%

%Appendix one text goes here.

% you can choose not to have a title for an appendix
% if you want by leaving the argument blank
%\section{}
%Appendix two text goes here.

% use section* for acknowledgment
%\section*{Acknowledgment}

%The authors would like to thank...

% Can use something like this to put references on a page
% by themselves when using endfloat and the captionsoff option.
\ifCLASSOPTIONcaptionsoff
  \newpage
\fi

% biography section
% 
% If you have an EPS/PDF photo (graphicx package needed) extra braces are
% needed around the contents of the optional argument to biography to prevent
% the LaTeX parser from getting confused when it sees the complicated
% \includegraphics command within an optional argument. (You could create
% your own custom macro containing the \includegraphics command to make things
% simpler here.)
%\begin{IEEEbiography}[{\includegraphics[width=1in,height=1.25in,clip,keepaspectratio]{mshell}}]{Michael Shell}
% or if you just want to reserve a space for a photo:

%\begin{IEEEbiography}{Michael Shell}
%Biography text here.
%\end{IEEEbiography}

% if you will not have a photo at all:
%\begin{IEEEbiographynophoto}{John Doe}
%Biography text here.
%\end{IEEEbiographynophoto}

% insert where needed to balance the two columns on the last page with
% biographies
%\newpage

%\begin{IEEEbiographynophoto}{Jane Doe}
%Biography text here.
%\end{IEEEbiographynophoto}

% You can push biographies down or up by placing
% a \vfill before or after them. The appropriate
% use of \vfill depends on what kind of text is
% on the last page and whether or not the columns
% are being equalized.

\bibliographystyle{IEEEtran}
\bibliography{reference}

% Generated by IEEEtran.bst, version: 1.14 (2015/08/26)
\begin{thebibliography}{10}
\providecommand{\url}[1]{#1}
\csname url@samestyle\endcsname
\providecommand{\newblock}{\relax}
\providecommand{\bibinfo}[2]{#2}
\providecommand{\BIBentrySTDinterwordspacing}{\spaceskip=0pt\relax}
\providecommand{\BIBentryALTinterwordstretchfactor}{4}
\providecommand{\BIBentryALTinterwordspacing}{\spaceskip=\fontdimen2\font plus
\BIBentryALTinterwordstretchfactor\fontdimen3\font minus
  \fontdimen4\font\relax}
\providecommand{\BIBforeignlanguage}[2]{{%
\expandafter\ifx\csname l@#1\endcsname\relax
\typeout{** WARNING: IEEEtran.bst: No hyphenation pattern has been}%
\typeout{** loaded for the language `#1'. Using the pattern for}%
\typeout{** the default language instead.}%
\else
\language=\csname l@#1\endcsname
\fi
#2}}
\providecommand{\BIBdecl}{\relax}
\BIBdecl

\bibitem{zhang2019marine}
B.~Zhang, E.~J. Matchinski, B.~Chen, X.~Ye, L.~Jing, and K.~Lee, ``Marine oil
  spills—oil pollution, sources and effects,'' \emph{World seas: an
  environmental evaluation}, pp. 391--406, 2019.

\bibitem{hjermann2007fish}
D.~{\O}. Hjermann, A.~Melsom, G.~E. Dings{\o}r, J.~M. Durant, A.~M. Eikeset,
  L.~P. R{\o}ed, G.~Ottersen, G.~Storvik, and N.~C. Stenseth, ``Fish and oil in
  the lofoten--barents sea system: synoptic review of the effect of oil spills
  on fish populations,'' \emph{Marine Ecology Progress Series}, vol. 339, pp.
  283--299, 2007.

\bibitem{farrington2014oil}
J.~W. Farrington, ``Oil pollution in the marine environment ii: fates and
  effects of oil spills,'' \emph{Environment: Science and Policy for
  Sustainable Development}, vol.~56, no.~4, pp. 16--31, 2014.

\bibitem{kingston2002long}
P.~F. Kingston, ``Long-term environmental impact of oil spills,'' \emph{Spill
  Science \& Technology Bulletin}, vol.~7, no. 1-2, pp. 53--61, 2002.

\bibitem{krupp2008impact}
F.~Krupp and A.~H. Abuzinada, ``Impact of oil pollution and increased sea
  surface temperatures on marine ecosystems and biota in the gulf,''
  \emph{Protecting the Gulf’s marine ecosystems from pollution}, pp. 45--56,
  2008.

\bibitem{soukissian2016satellite}
T.~Soukissian, F.~Karathanasi, and P.~Axaopoulos, ``Satellite-based offshore
  wind resource assessment in the mediterranean sea,'' \emph{IEEE Journal of
  Oceanic Engineering}, vol.~42, no.~1, pp. 73--86, 2016.

\bibitem{li2017wind}
H.~Li, J.~Wu, W.~Perrie, and Y.~He, ``Wind speed retrieval from hybrid-pol
  compact polarization synthetic aperture radar images,'' \emph{IEEE Journal of
  Oceanic Engineering}, vol.~43, no.~3, pp. 713--724, 2017.

\bibitem{solberg2007oil}
A.~H. Solberg, C.~Brekke, and P.~O. Husoy, ``Oil spill detection in radarsat
  and envisat sar images,'' \emph{IEEE Transactions on Geoscience and Remote
  Sensing}, vol.~45, no.~3, pp. 746--755, 2007.

\bibitem{brekke2005oil}
C.~Brekke and A.~H. Solberg, ``Oil spill detection by satellite remote
  sensing,'' \emph{Remote sensing of environment}, vol.~95, no.~1, pp. 1--13,
  2005.

\bibitem{velotto2016first}
D.~Velotto, C.~Bentes, B.~Tings, and S.~Lehner, ``First comparison of
  sentinel-1 and terrasar-x data in the framework of maritime targets
  detection: South italy case,'' \emph{IEEE Journal of Oceanic Engineering},
  vol.~41, no.~4, pp. 993--1006, 2016.

\bibitem{yin2014modified}
J.~Yin and J.~Yang, ``A modified level set approach for segmentation of
  multiband polarimetric sar images,'' \emph{IEEE Transactions on Geoscience
  and Remote Sensing}, vol.~52, no.~11, pp. 7222--7232, 2014.

\bibitem{salberg2014oil}
A.-B. Salberg, {\O}.~Rudjord, and A.~H.~S. Solberg, ``Oil spill detection in
  hybrid-polarimetric sar images,'' \emph{IEEE Transactions on Geoscience and
  Remote Sensing}, vol.~52, no.~10, pp. 6521--6533, 2014.

\bibitem{buono2016polarimetric}
A.~Buono, F.~Nunziata, M.~Migliaccio, and X.~Li, ``Polarimetric analysis of
  compact-polarimetry sar architectures for sea oil slick observation,''
  \emph{IEEE Transactions on Geoscience and Remote Sensing}, vol.~54, no.~10,
  pp. 5862--5874, 2016.

\bibitem{velotto2011dual}
D.~Velotto, M.~Migliaccio, F.~Nunziata, and S.~Lehner, ``Dual-polarized
  terrasar-x data for oil-spill observation,'' \emph{IEEE Transactions on
  geoscience and remote sensing}, vol.~49, no.~12, pp. 4751--4762, 2011.

\bibitem{espeseth2017analysis}
M.~M. Espeseth, S.~Skrunes, C.~E. Jones, C.~Brekke, B.~Holt, and A.~P.
  Doulgeris, ``Analysis of evolving oil spills in full-polarimetric and
  hybrid-polarity sar,'' \emph{IEEE Transactions on Geoscience and Remote
  Sensing}, vol.~55, no.~7, pp. 4190--4210, 2017.

\bibitem{minchew2012polarimetric}
B.~Minchew, C.~E. Jones, and B.~Holt, ``Polarimetric analysis of backscatter
  from the deepwater horizon oil spill using l-band synthetic aperture radar,''
  \emph{IEEE Transactions on Geoscience and Remote Sensing}, vol.~50, no.~10,
  pp. 3812--3830, 2012.

\bibitem{collins2015use}
M.~J. Collins, M.~Denbina, B.~Minchew, C.~E. Jones, and B.~Holt, ``On the use
  of simulated airborne compact polarimetric sar for characterizing oil--water
  mixing of the deepwater horizon oil spill,'' \emph{IEEE Journal of Selected
  Topics in Applied Earth Observations and Remote Sensing}, vol.~8, no.~3, pp.
  1062--1077, 2015.

\bibitem{gemme2018automatic}
L.~Gemme and S.~G. Dellepiane, ``An automatic data-driven method for sar image
  segmentation in sea surface analysis,'' \emph{IEEE Transactions on Geoscience
  and Remote Sensing}, vol.~56, no.~5, pp. 2633--2646, 2018.

\bibitem{lupidi2017fast}
A.~Lupidi, D.~Staglian{\`o}, M.~Martorella, and F.~Berizzi, ``Fast detection of
  oil spills and ships using sar images,'' \emph{Remote Sensing}, vol.~9,
  no.~3, p. 230, 2017.

\bibitem{marques2011sar}
R.~C.~P. Marques, F.~N. Medeiros, and J.~S. Nobre, ``Sar image segmentation
  based on level set approach and $\{$$\backslash$cal G$\}$ \_a\^{} 0 model,''
  \emph{IEEE transactions on pattern analysis and machine intelligence},
  vol.~34, no.~10, pp. 2046--2057, 2011.

\bibitem{xu2017level}
M.~Xu, Y.~Yu, F.~Chen, X.~Jiang, P.~Ren, and E.~Yang, ``Level sets with one dot
  fuzzy initialization for marine oil spill segmentation,'' \emph{OCEANS
  2017-Aberdeen}, pp. 1--4, 2017.

\bibitem{xia2015meaningful}
G.-S. Xia, G.~Liu, W.~Yang, and L.~Zhang, ``Meaningful object segmentation from
  sar images via a multiscale nonlocal active contour model,'' \emph{IEEE
  Transactions on Geoscience and Remote Sensing}, vol.~54, no.~3, pp.
  1860--1873, 2015.

\bibitem{chen2018segmenting}
F.~Chen, H.~Zhou, C.~Grecos, and P.~Ren, ``Segmenting oil spills from blurry
  images based on alternating direction method of multipliers,'' \emph{IEEE
  Journal of Selected Topics in Applied Earth Observations and Remote Sensing},
  vol.~11, no.~6, pp. 1858--1873, 2018.

\bibitem{jing2011novel}
Y.~Jing, J.~An, and Z.~Liu, ``A novel edge detection algorithm based on global
  minimization active contour model for oil slick infrared aerial image,''
  \emph{IEEE Transactions on Geoscience and Remote sensing}, vol.~49, no.~6,
  pp. 2005--2013, 2011.

\bibitem{chen2017level}
F.~Chen, X.~Yu, X.~Jiang, and P.~Ren, ``Level sets with self-guided filtering
  for marine oil spill segmentation,'' \emph{2017 IEEE International Geoscience
  and Remote Sensing Symposium (IGARSS)}, pp. 1772--1775, 2017.

\bibitem{ren2018energy}
P.~Ren, M.~Xu, Y.~Yu, F.~Chen, X.~Jiang, and E.~Yang, ``Energy minimization
  with one dot fuzzy initialization for marine oil spill segmentation,''
  \emph{IEEE Journal of Oceanic Engineering}, vol.~44, no.~4, pp. 1102--1115,
  2018.

\bibitem{bertacca2005farima}
M.~Bertacca, F.~Berizzi, and E.~D. Mese, ``A farima-based technique for oil
  slick and low-wind areas discrimination in sea sar imagery,'' \emph{IEEE
  transactions on geoscience and remote sensing}, vol.~43, no.~11, pp.
  2484--2493, 2005.

\bibitem{hinton2006reducing}
G.~E. Hinton and R.~R. Salakhutdinov, ``Reducing the dimensionality of data
  with neural networks,'' \emph{science}, vol. 313, no. 5786, pp. 504--507,
  2006.

\bibitem{minaee2021image}
S.~Minaee, Y.~Y. Boykov, F.~Porikli, A.~J. Plaza, N.~Kehtarnavaz, and
  D.~Terzopoulos, ``Image segmentation using deep learning: A survey,''
  \emph{IEEE transactions on pattern analysis and machine intelligence}, 2021.

\bibitem{krizhevsky2017imagenet}
A.~Krizhevsky, I.~Sutskever, and G.~E. Hinton, ``Imagenet classification with
  deep convolutional neural networks,'' \emph{Communications of the ACM},
  vol.~60, no.~6, pp. 84--90, 2017.

\bibitem{bengio2009learning}
Y.~Bengio \emph{et~al.}, ``Learning deep architectures for ai,''
  \emph{Foundations and trends{\textregistered} in Machine Learning}, vol.~2,
  no.~1, pp. 1--127, 2009.

\bibitem{ulaby2014microwave}
F.~Ulaby and D.~Long, \emph{Microwave radar and radiometric remote
  sensing}.\hskip 1em plus 0.5em minus 0.4em\relax University of Michigan
  Press, 2014, vol.~4, no.~5.

\bibitem{oliver2004understanding}
C.~Oliver and S.~Quegan, \emph{Understanding synthetic aperture radar
  images}.\hskip 1em plus 0.5em minus 0.4em\relax SciTech Publishing, 2004.

\bibitem{alpers2017oil}
W.~Alpers, B.~Holt, and K.~Zeng, ``Oil spill detection by imaging radars:
  Challenges and pitfalls,'' \emph{Remote sensing of environment}, vol. 201,
  pp. 133--147, 2017.

\bibitem{chen2023dgnet}
F.~Chen, H.~Balzter, F.~Zhou, P.~Ren, and H.~Zhou, ``Dgnet: Distribution guided
  efficient learning for oil spill image segmentation,'' \emph{IEEE
  Transactions on Geoscience and Remote Sensing}, vol.~61, pp. 1--17, 2023.

\bibitem{pathak2016context}
D.~Pathak, P.~Krahenbuhl, J.~Donahue, T.~Darrell, and A.~A. Efros, ``Context
  encoders: Feature learning by inpainting,'' in \emph{Proceedings of the IEEE
  conference on computer vision and pattern recognition}, 2016, pp. 2536--2544.

\bibitem{kingma2014adam}
D.~P. Kingma and J.~Ba, ``Adam: A method for stochastic optimization,''
  \emph{arXiv preprint arXiv:1412.6980}, 2014.

\bibitem{nieto2018two}
M.~Nieto-Hidalgo, A.-J. Gallego, P.~Gil, and A.~Pertusa, ``Two-stage
  convolutional neural network for ship and spill detection using slar
  images,'' \emph{IEEE Transactions on geoscience and remote sensing}, vol.~56,
  no.~9, pp. 5217--5230, 2018.

\bibitem{garcia2017review}
A.~Garcia-Garcia, S.~Orts-Escolano, S.~Oprea, V.~Villena-Martinez, and
  J.~Garcia-Rodriguez, ``A review on deep learning techniques applied to
  semantic segmentation,'' \emph{arXiv preprint arXiv:1704.06857}, 2017.

\bibitem{cantorna2019oil}
D.~Cantorna, C.~Dafonte, A.~Iglesias, and B.~Arcay, ``Oil spill segmentation in
  sar images using convolutional neural networks. a comparative analysis with
  clustering and logistic regression algorithms,'' \emph{Applied Soft
  Computing}, vol.~84, p. 105716, 2019.

\bibitem{muruganandham2016semantic}
S.~Muruganandham, ``Semantic segmentation of satellite images using deep
  learning,'' 2016.

\bibitem{li2011level}
C.~Li, R.~Huang, Z.~Ding, J.~C. Gatenby, D.~N. Metaxas, and J.~C. Gore, ``A
  level set method for image segmentation in the presence of intensity
  inhomogeneities with application to mri,'' \emph{IEEE transactions on image
  processing}, vol.~20, no.~7, pp. 2007--2016, 2011.

\bibitem{zhang2015level}
K.~Zhang, L.~Zhang, K.-M. Lam, and D.~Zhang, ``A level set approach to image
  segmentation with intensity inhomogeneity,'' \emph{IEEE transactions on
  cybernetics}, vol.~46, no.~2, pp. 546--557, 2015.

\bibitem{goodfellow2020generative}
I.~Goodfellow, J.~Pouget-Abadie, M.~Mirza, B.~Xu, D.~Warde-Farley, S.~Ozair,
  A.~Courville, and Y.~Bengio, ``Generative adversarial networks,''
  \emph{Communications of the ACM}, vol.~63, no.~11, pp. 139--144, 2020.

\bibitem{yu2018oil}
X.~Yu, H.~Zhang, C.~Luo, H.~Qi, and P.~Ren, ``Oil spill segmentation via
  adversarial $ f $-divergence learning,'' \emph{IEEE Transactions on
  Geoscience and Remote Sensing}, vol.~56, no.~9, pp. 4973--4988, 2018.

\bibitem{zhao2018icnet}
H.~Zhao, X.~Qi, X.~Shen, J.~Shi, and J.~Jia, ``Icnet for real-time semantic
  segmentation on high-resolution images,'' in \emph{Proceedings of the
  European conference on computer vision (ECCV)}, 2018, pp. 405--420.

\bibitem{ronneberger2015u}
O.~Ronneberger, P.~Fischer, and T.~Brox, ``U-net: Convolutional networks for
  biomedical image segmentation,'' in \emph{International Conference on Medical
  image computing and computer-assisted intervention}.\hskip 1em plus 0.5em
  minus 0.4em\relax Springer, 2015, pp. 234--241.

\bibitem{long2015fully}
J.~Long, E.~Shelhamer, and T.~Darrell, ``Fully convolutional networks for
  semantic segmentation,'' in \emph{Proceedings of the IEEE conference on
  computer vision and pattern recognition}, 2015, pp. 3431--3440.

\end{thebibliography}

\end{document}